\documentclass[letterpaper]{article}
\usepackage{aaai}
\usepackage{times}
\usepackage{helvet}
\usepackage{courier}
\usepackage[hyphens]{url}  
\usepackage{graphicx} 
\usepackage{natbib}  
\usepackage{caption} 

\usepackage{bm}
\usepackage{amsmath}
\usepackage{multirow}
\usepackage{caption}
\usepackage{subfigure}

\begin{document}
%
\title{Adaptive Label Smoothing for Out-of-Distribution Detection}
\author{Mingle Xu\textsuperscript{\rm 1},
    Jaehwan Lee\textsuperscript{\rm 1},
    Sook Yoon\textsuperscript{\rm 2},
    Dong Sun Park\textsuperscript{\rm 1}\\
1 Electronic Engineerring, Jeonbuk National University, South Korea \\
2 Department of Computer Engineering, Mokpo National University, South Korea \\
}
\maketitle
\begin{abstract}
\begin{quote}
Out-of-distribution (OOD) detection, which aims to distinguish unknown classes from known classes, has received increasing attention recently. A main challenge within is the unavailable of samples from the unknown classes in the training process, and an effective strategy is to improve the performance for known classes. Using beneficial strategies such as data augmentation and longer training is thus a way to improve OOD detection. However, label smoothing, an effective method for classifying known classes, degrades the performance of OOD detection, and this phenomenon is under exploration. In this paper, we first analyze that the limited and predefined learning target in label smoothing results in the smaller maximal probability and logit, which further leads to worse OOD detection performance. To mitigate this issue, we then propose a novel regularization method, called adaptive label smoothing (ALS), and the core is to push the non-true classes to have same probabilities whereas the maximal probability is neither fixed nor limited. Extensive experimental results in six datasets with two backbones suggest that ALS contributes to classifying known samples and discerning unknown samples with clear margins. Our code will be available to the public.
\end{quote}
\end{abstract}

\section{Introduction}
Although deep learning has achieved significant improvement in the last decade, many deep learning methods elusively embrace the i.i.d. assumption \cite{vapnik1998statistical}, that training and test data are sampled independently from identical distributions and generally suffer in many real-world applications where the assumption is not satisfied often. A conflicting scenario is that new semantic classes appear in test \cite{scheirer2011meta, scheirer2012toward}. For example, collecting data for all classes is difficult and expensive considering the rate of natural occurrence and new things may occur \cite{meng2023known}. In this scenario, previous methods persist to classify the new classes into one of the training classes and are thus neither reliable nor safe. Out-of-distribution (OOD) detection \cite{hendrycks2022baseline, geng2020recent, yang2024generalized} is adopted to facilitate this issue where a test input from new semantic classes is required to be discern as unknown.

A main challenge in OOD detection is the inaccessibility of those unknown classes in the training process, in which learning discriminative features is prohibitive \cite{dietterich2022familiarity}. The empirical result suggests that improving known performance contributes to distinguishing unknown from known \cite{vaze2022open, meng2023known}. Inspired by this observation, methods that benefit the known classification can be used, such as strong data augmentation and longer training \cite{vaze2022open}. However, as a simple yet effective method and widely adopted, label smoothing \cite{szegedy2016rethinking, zhang2021delving} did not show its superiority in OOD detection \cite{lee2020soft, vaze2022open}. Why it is inferior is underexplored.

\begin{figure*}[ht!]
\centering
    \includegraphics[width=0.16\textwidth]{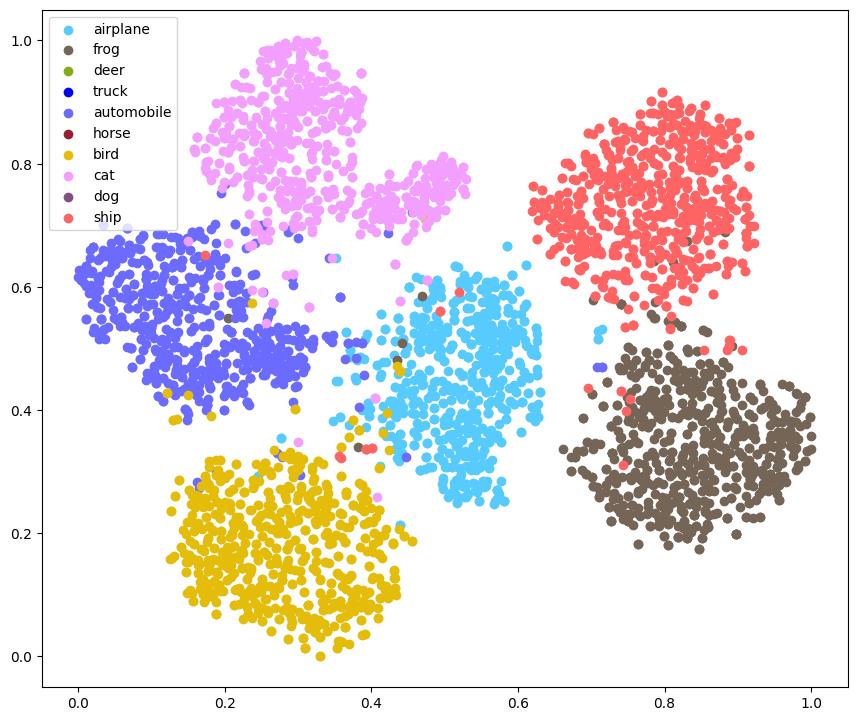}
    \includegraphics[width=0.16\textwidth]{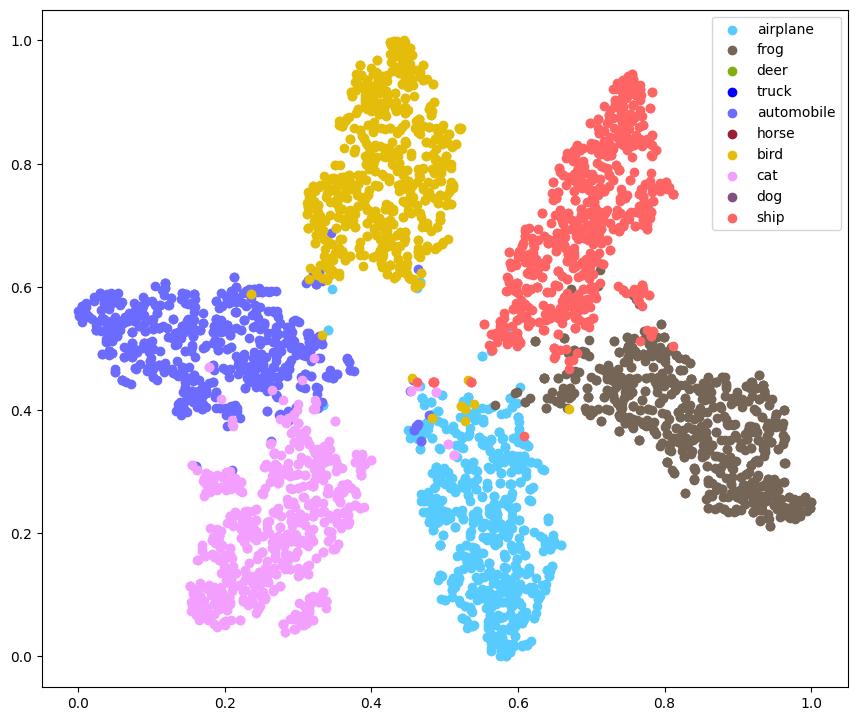}
    \includegraphics[width=0.16\textwidth]{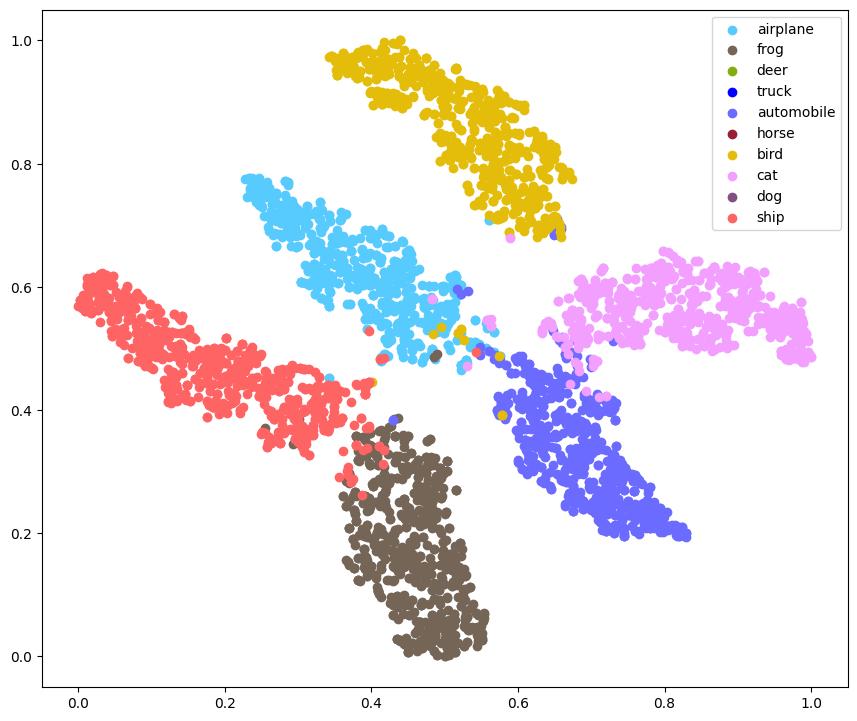}
    \includegraphics[width=0.16\textwidth]{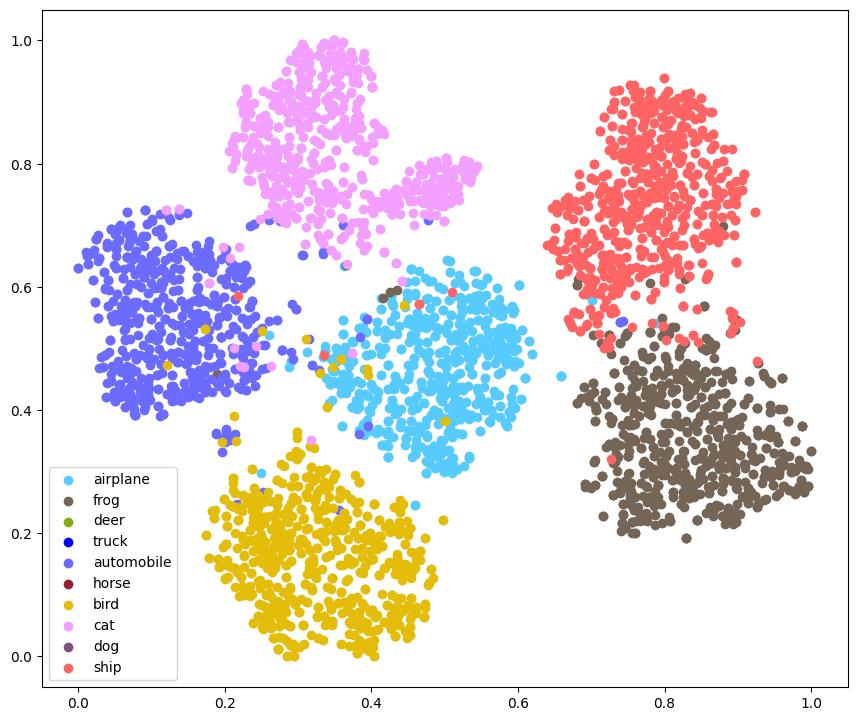}
    \includegraphics[width=0.16\textwidth]{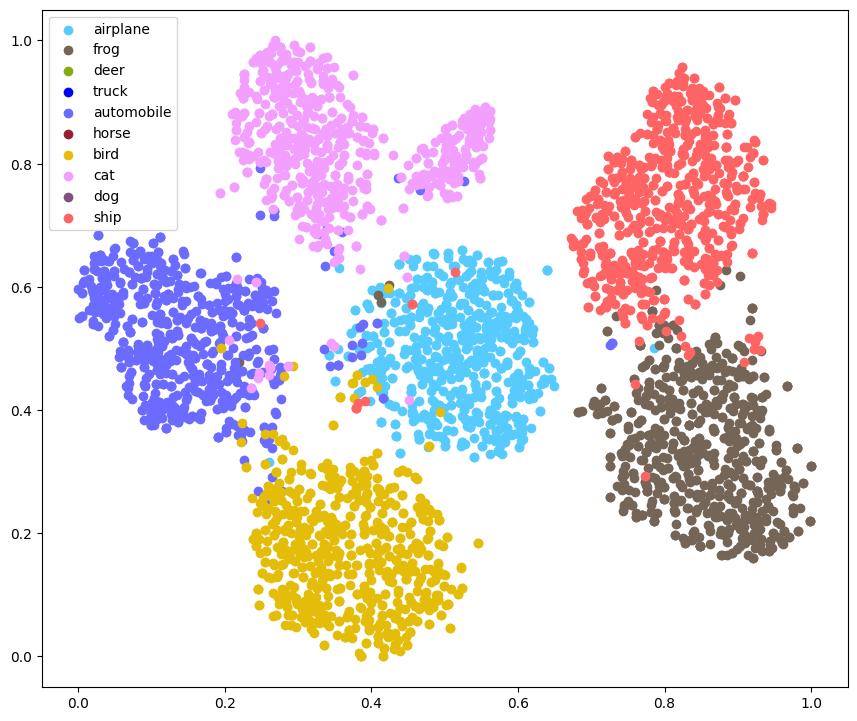}
    \includegraphics[width=0.16\textwidth]{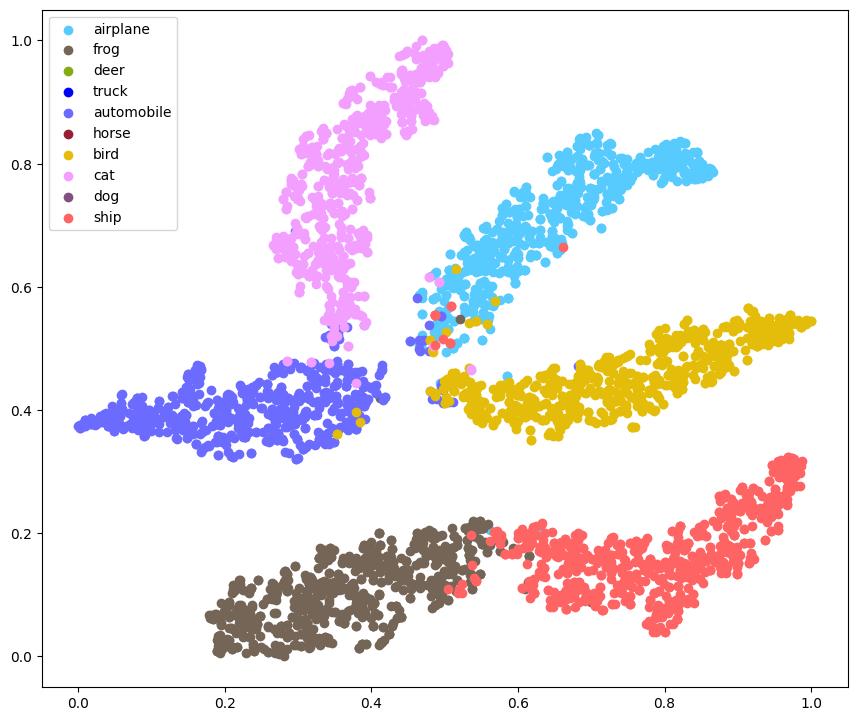}
    \\
    \includegraphics[width=0.16\textwidth]{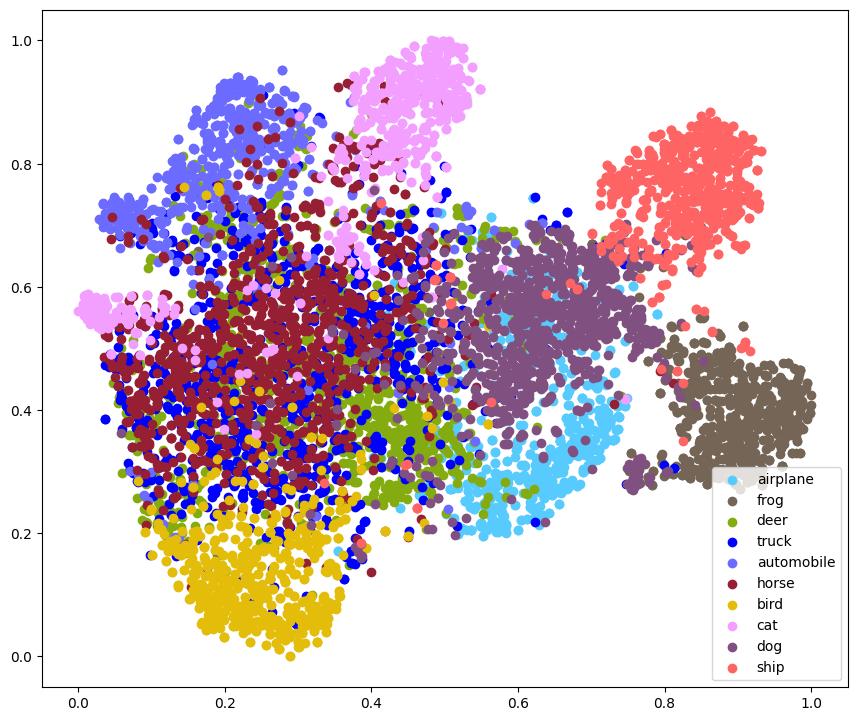}
    \includegraphics[width=0.16\textwidth]{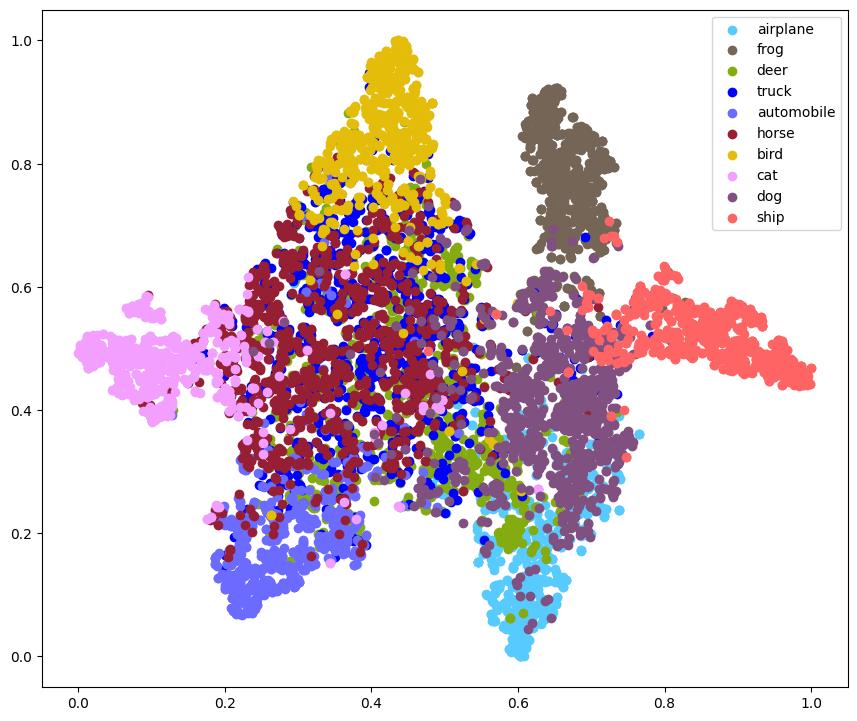}
    \includegraphics[width=0.16\textwidth]{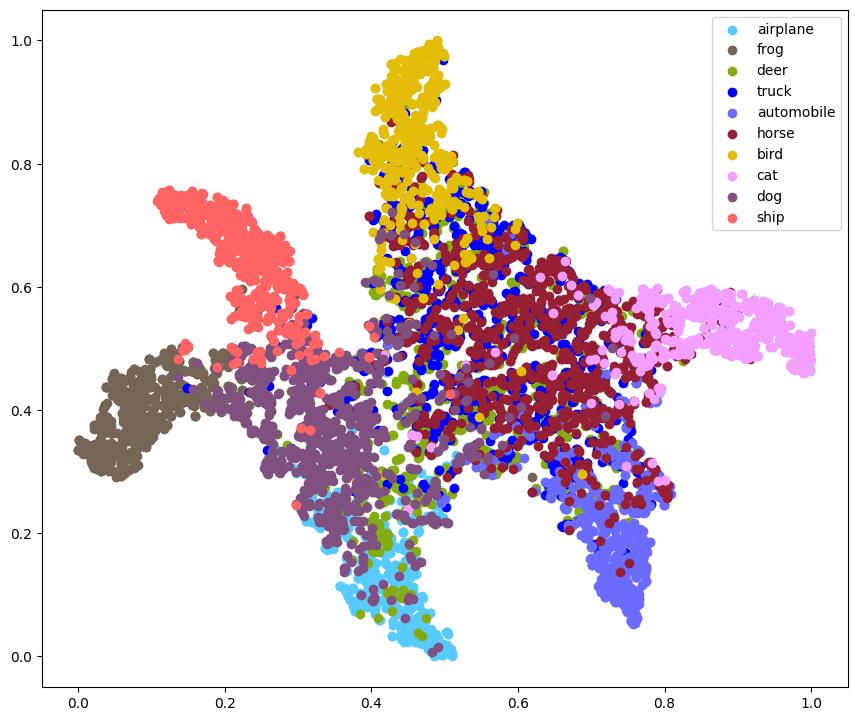}
    \includegraphics[width=0.16\textwidth]{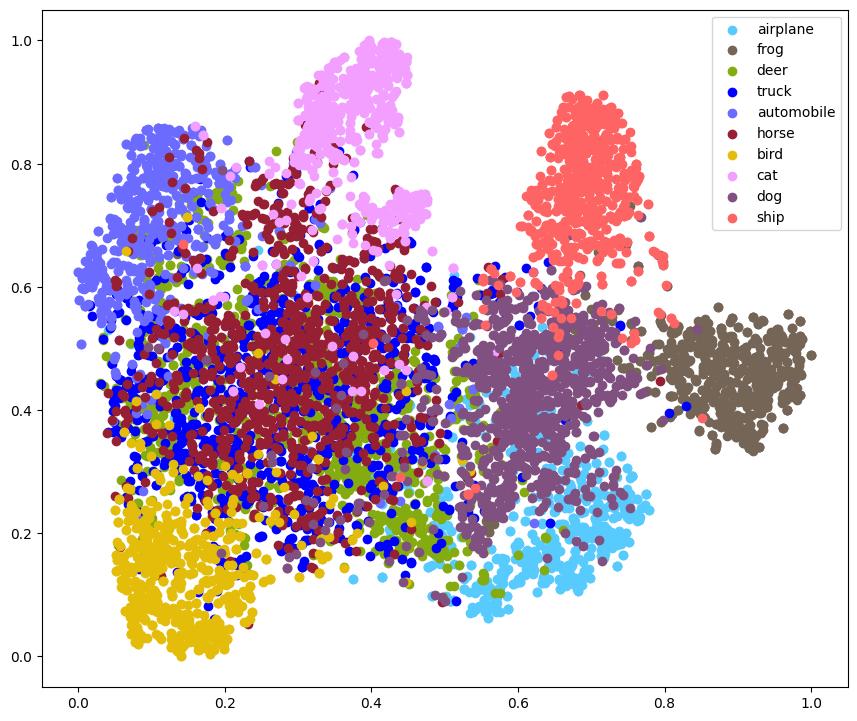}
    \includegraphics[width=0.16\textwidth]{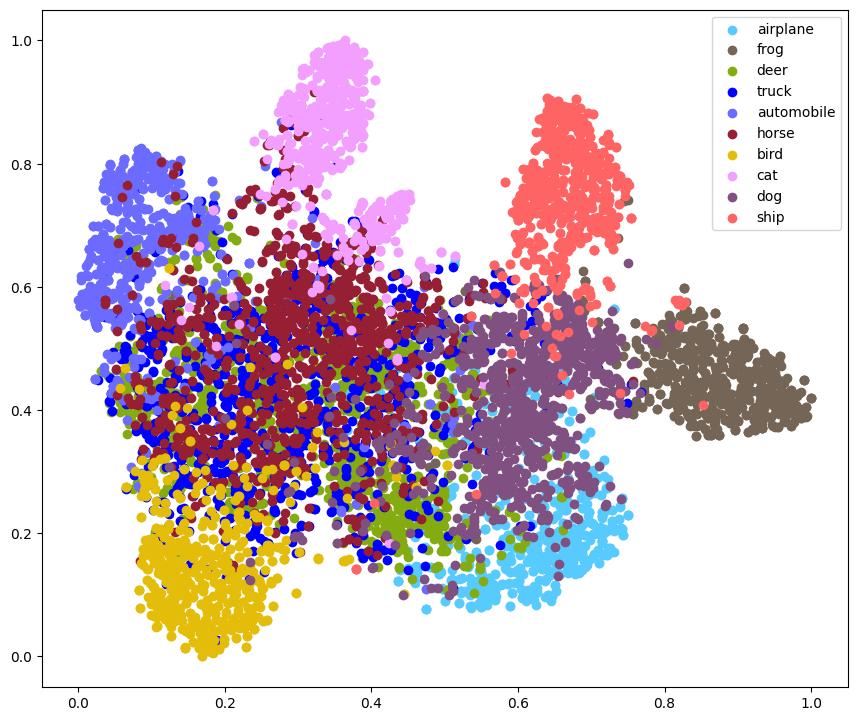}
    \includegraphics[width=0.16\textwidth]{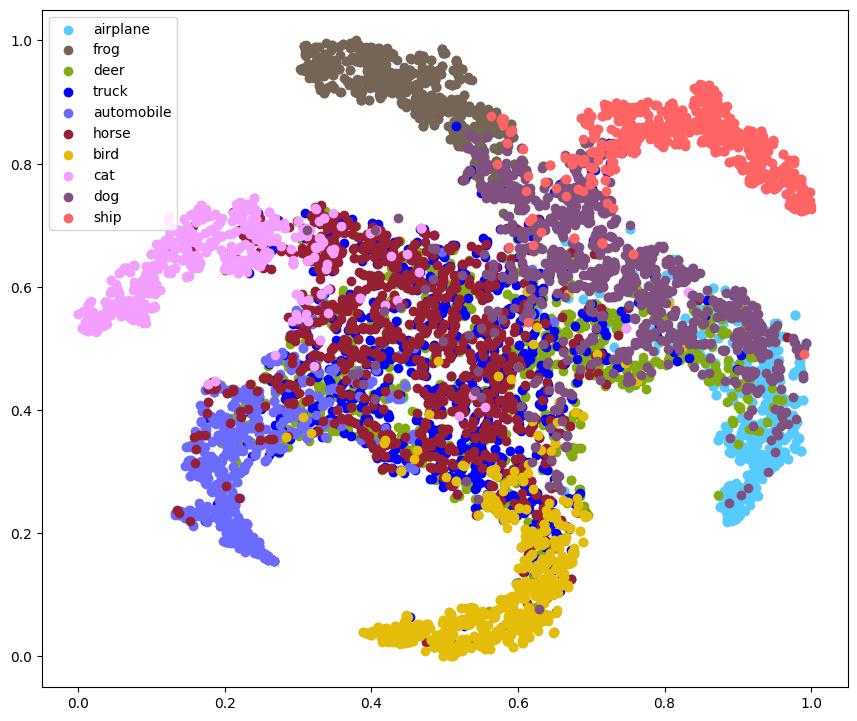}
    \\
    \includegraphics[width=0.16\textwidth]{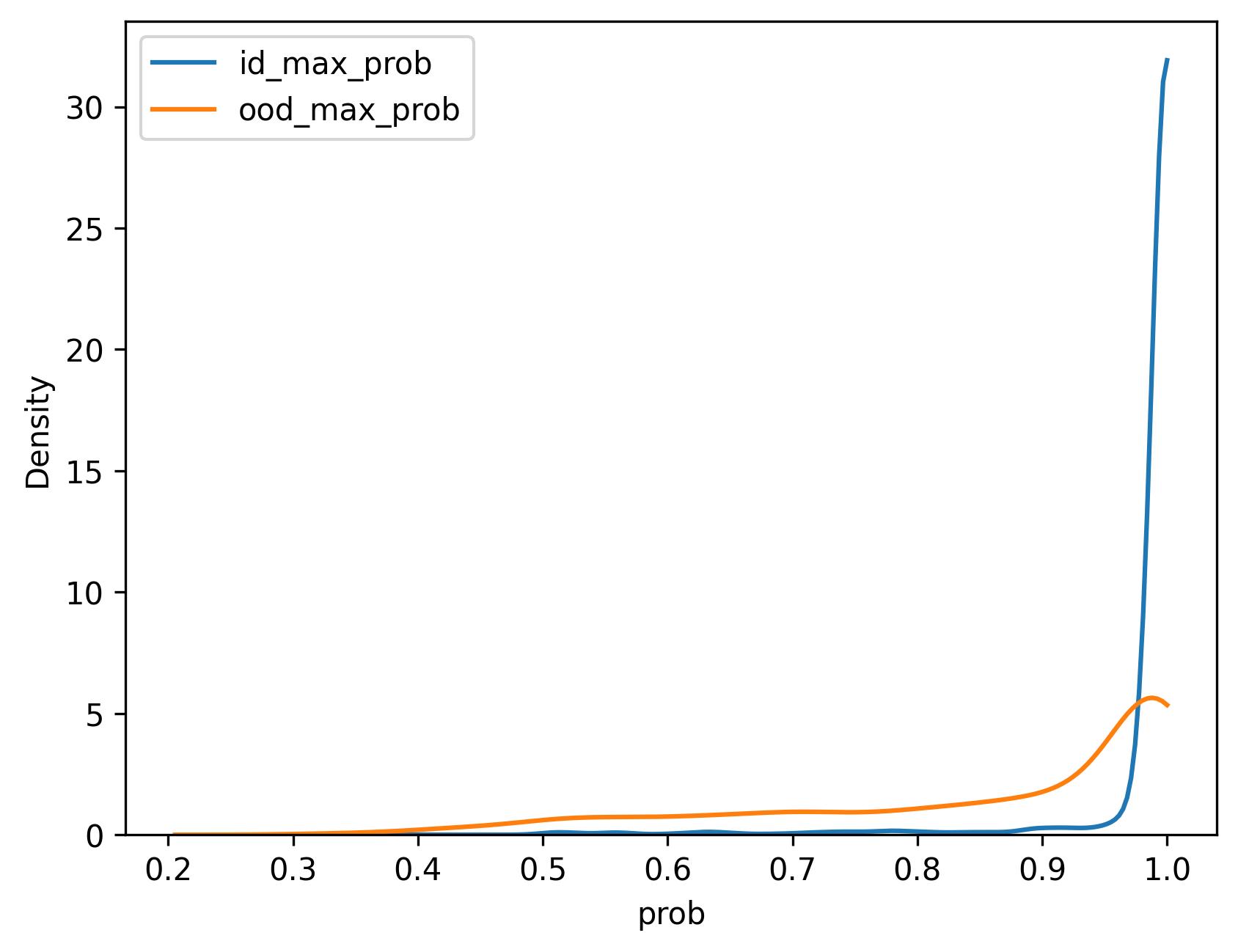}
    \includegraphics[width=0.16\textwidth]{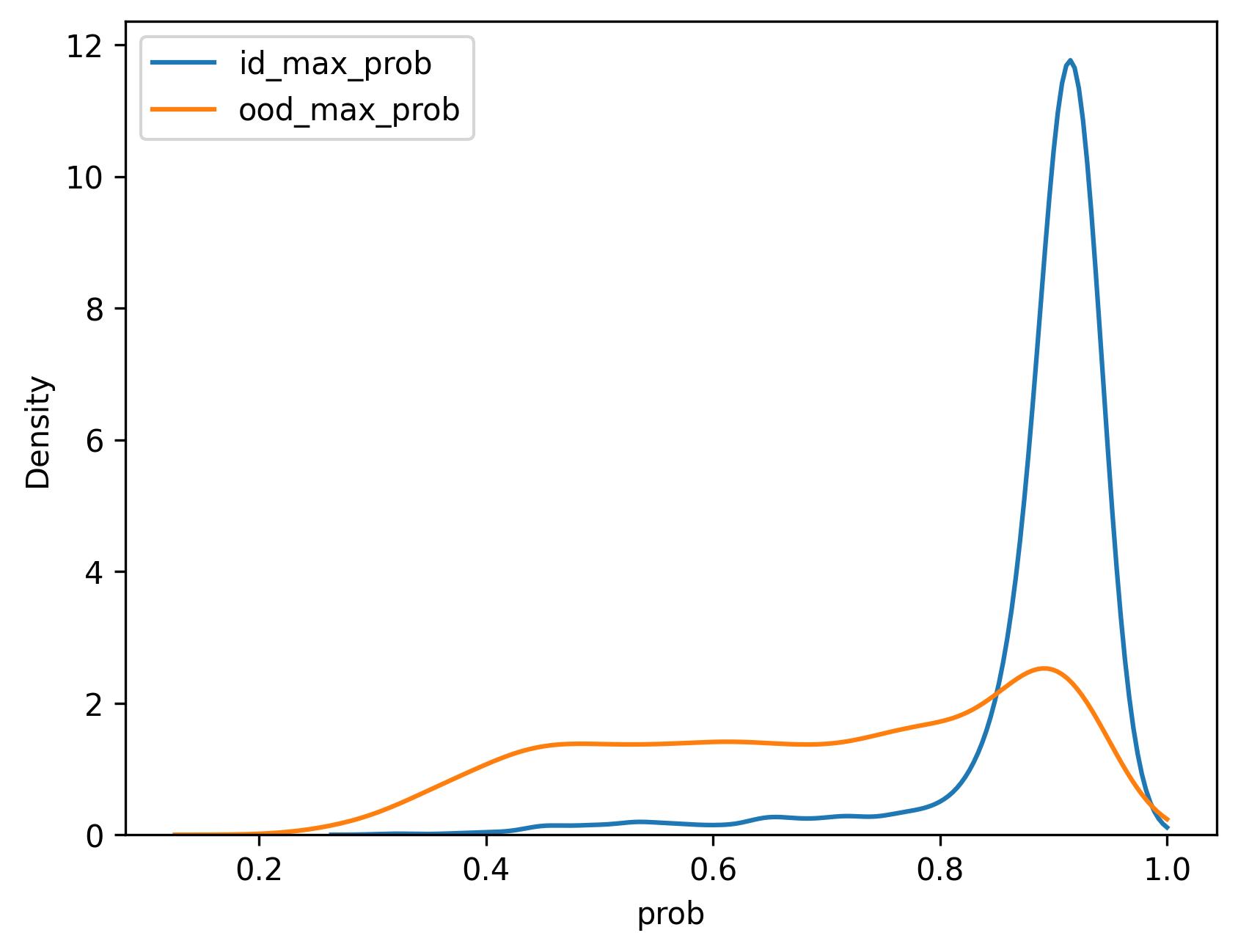}
    \includegraphics[width=0.16\textwidth]{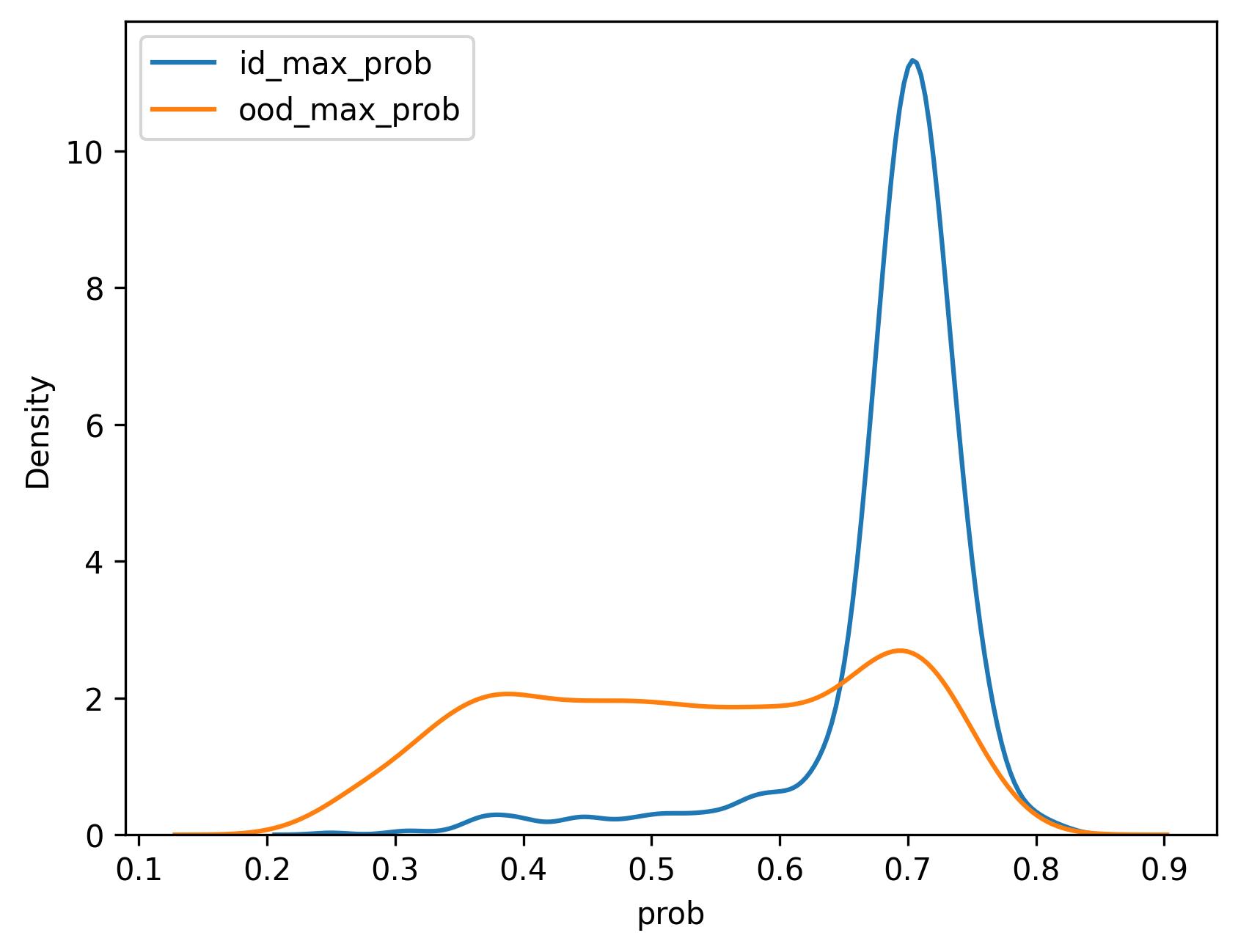}
    \includegraphics[width=0.16\textwidth]{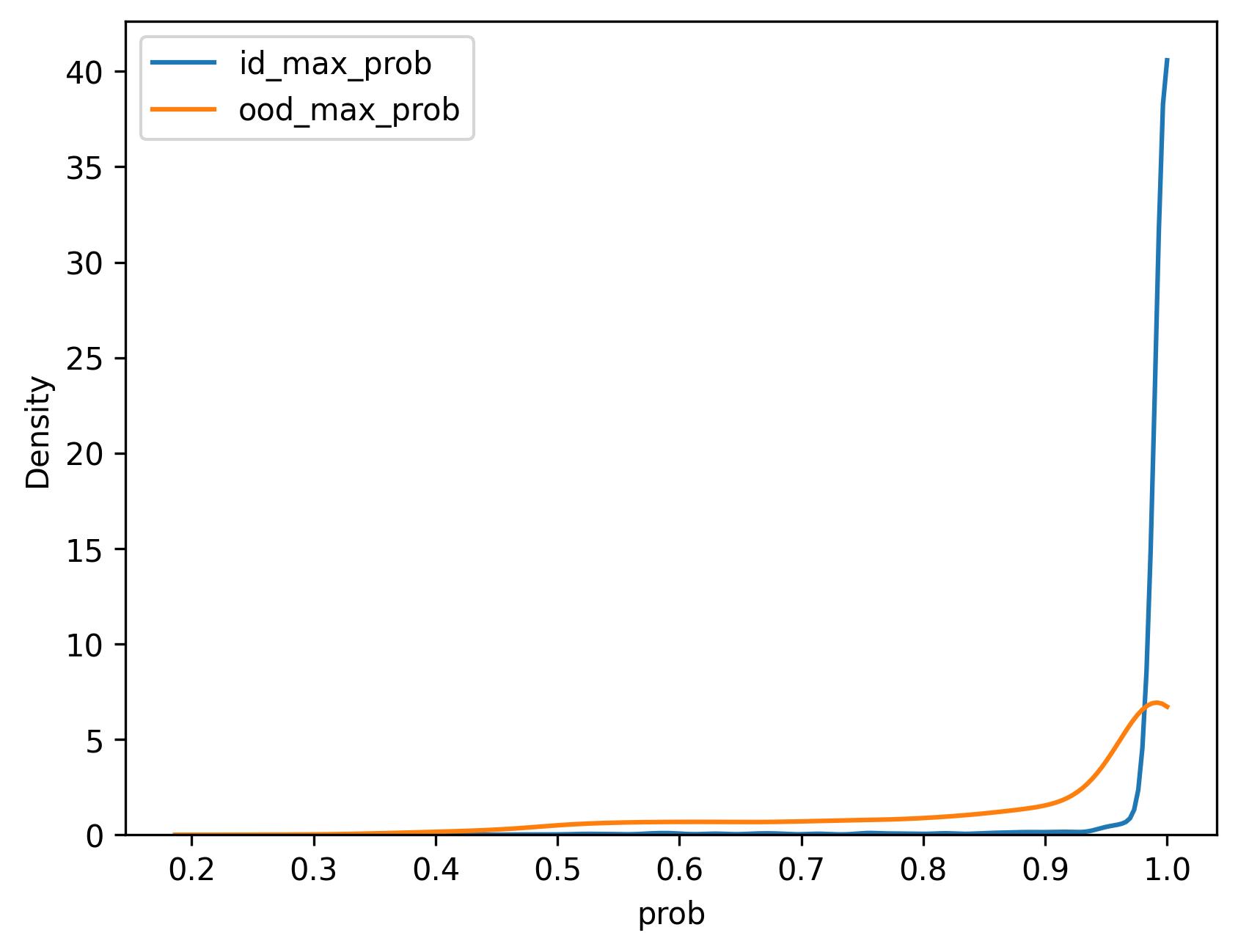}
    \includegraphics[width=0.16\textwidth]{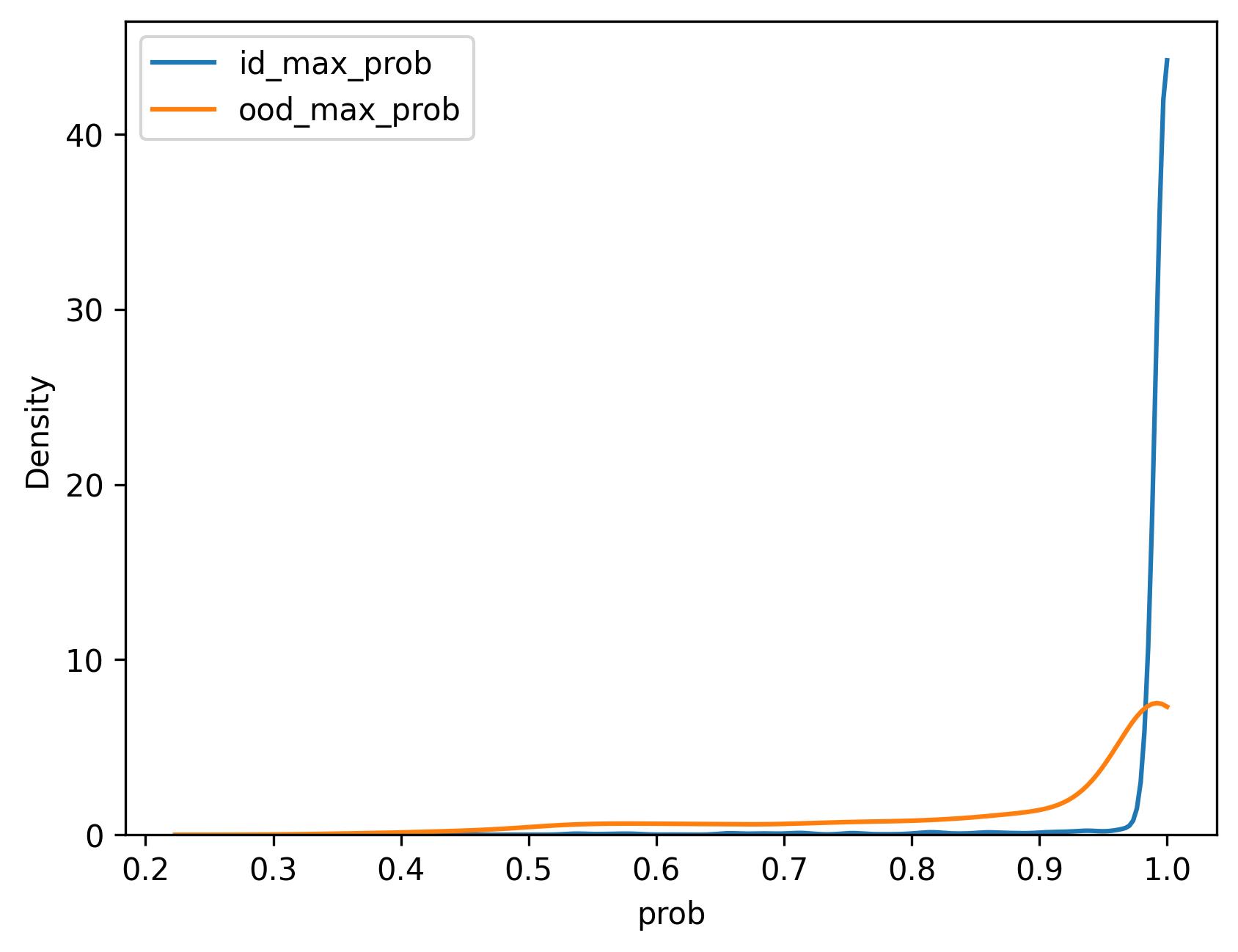}
    \includegraphics[width=0.16\textwidth]{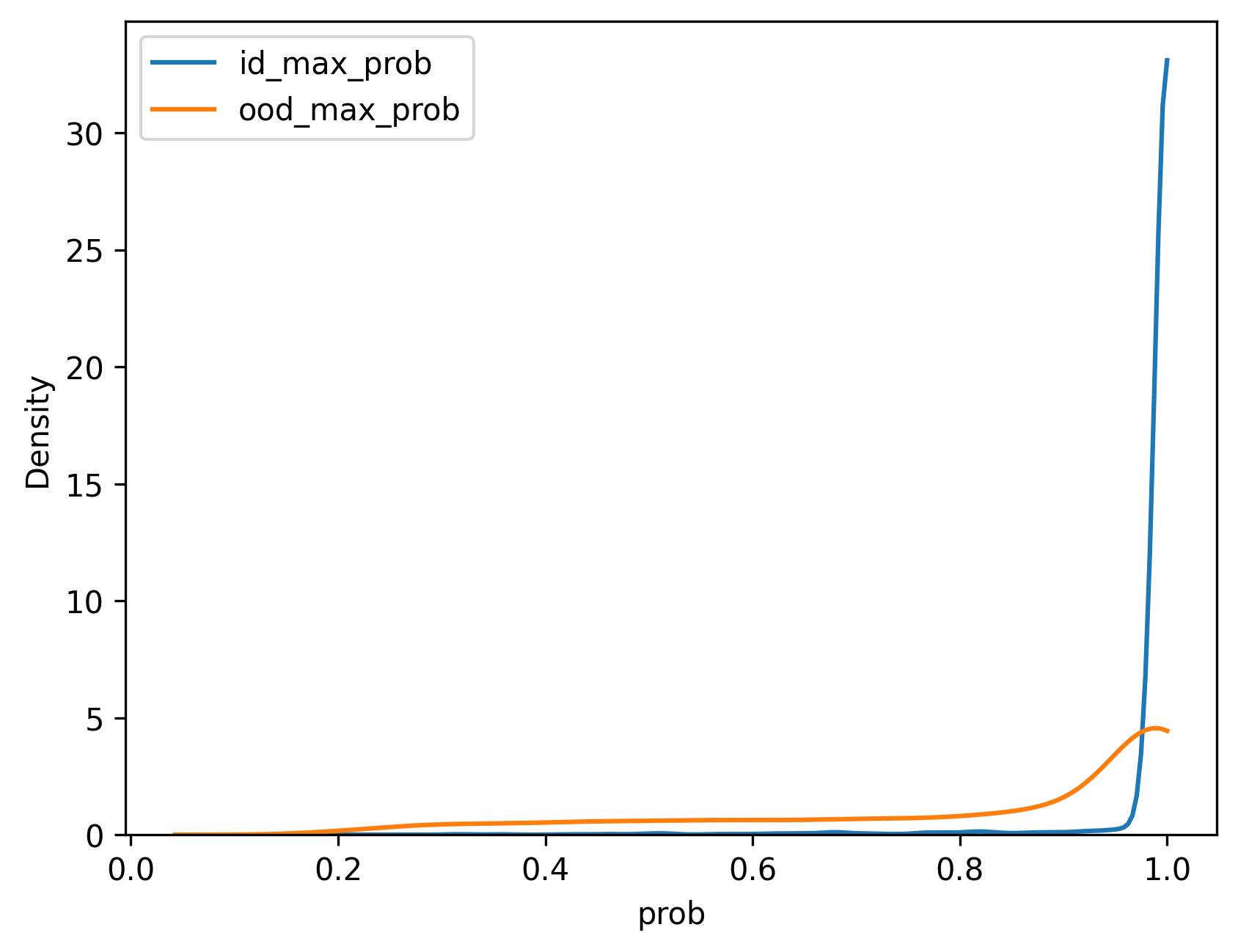}
    \\
    \subfigure[baseline]{\includegraphics[width=0.16\textwidth]{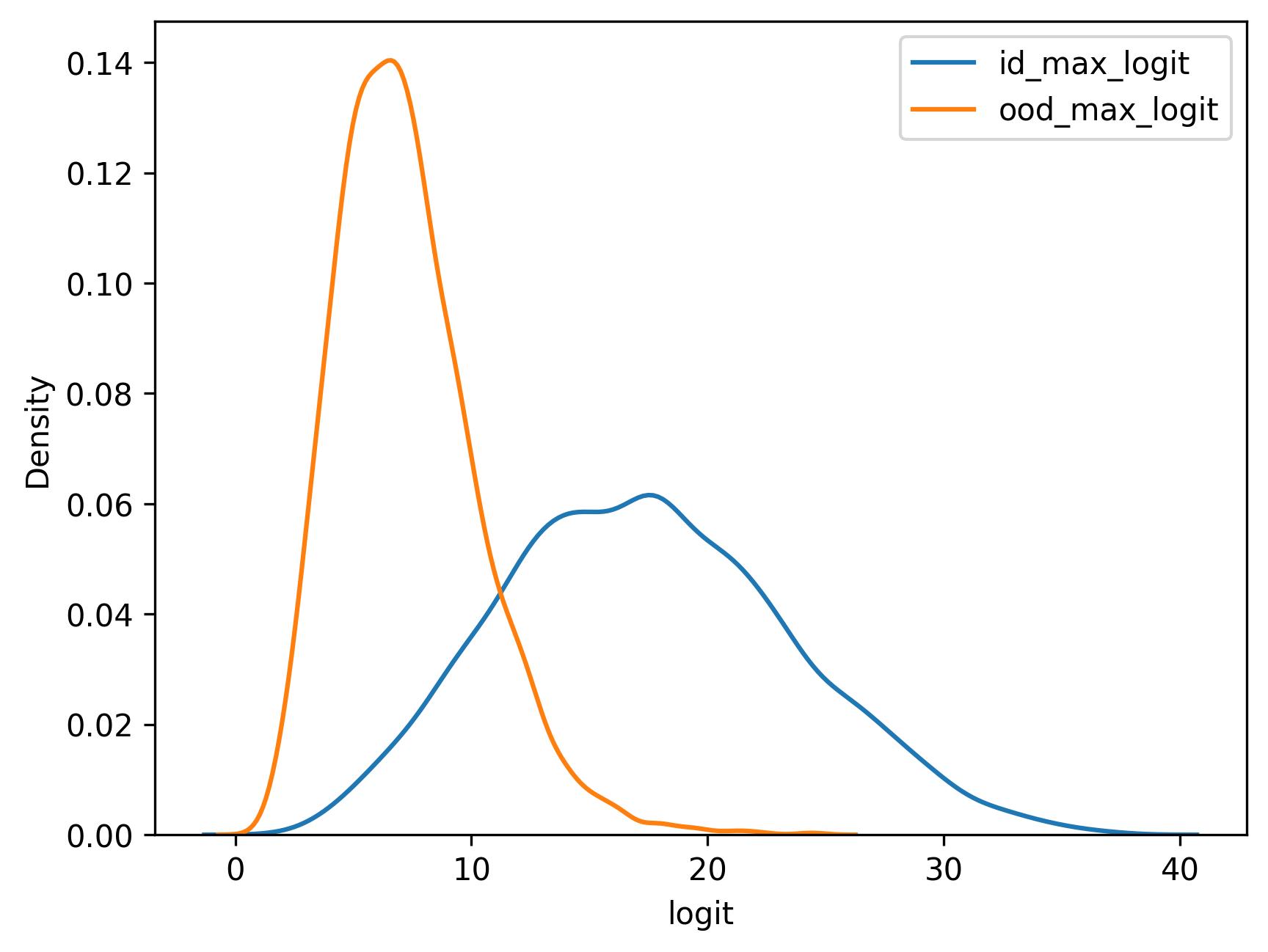}}
    \subfigure[LS 0.1]{\includegraphics[width=0.16\textwidth]{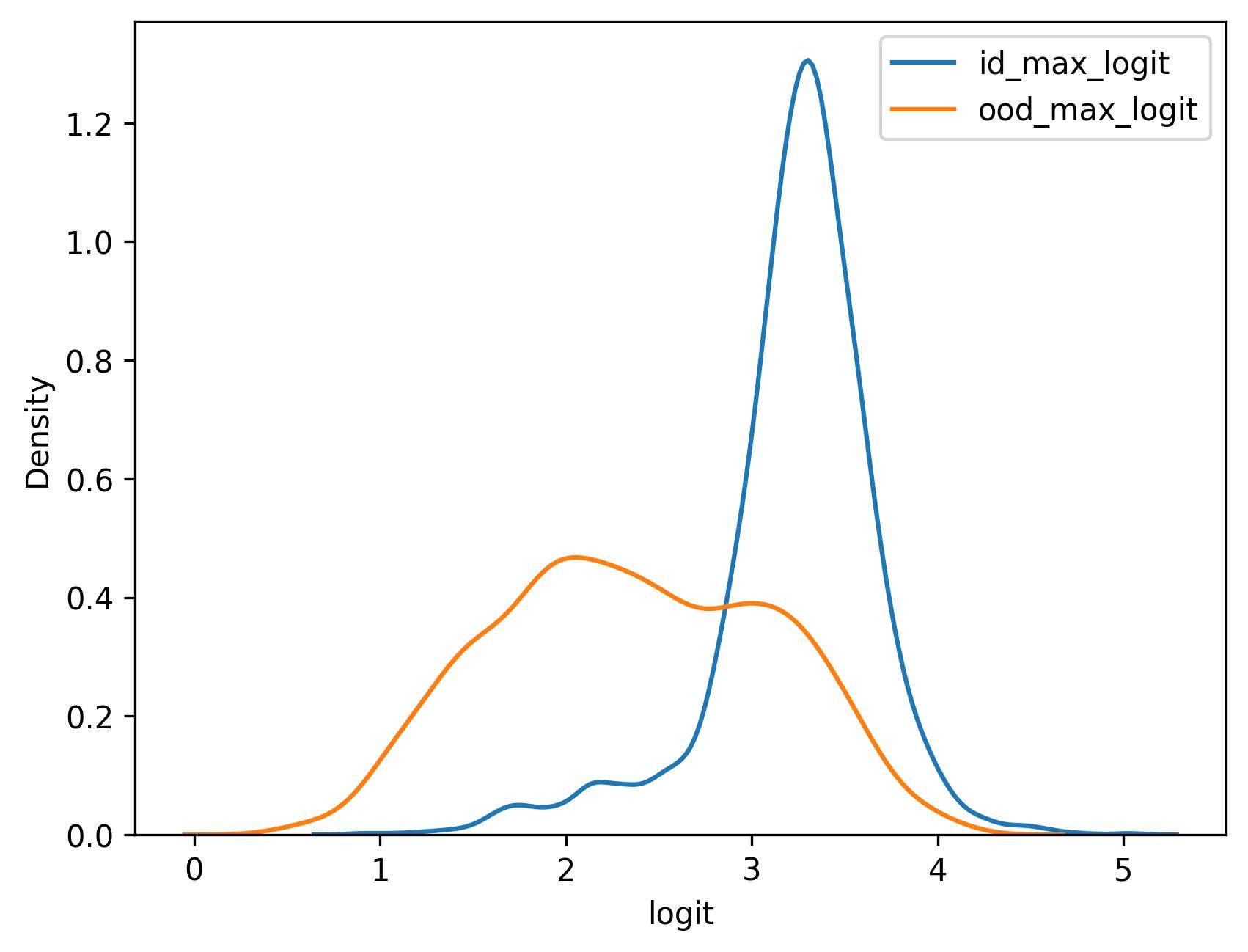}}
    \subfigure[LS 0.3]{\includegraphics[width=0.16\textwidth]{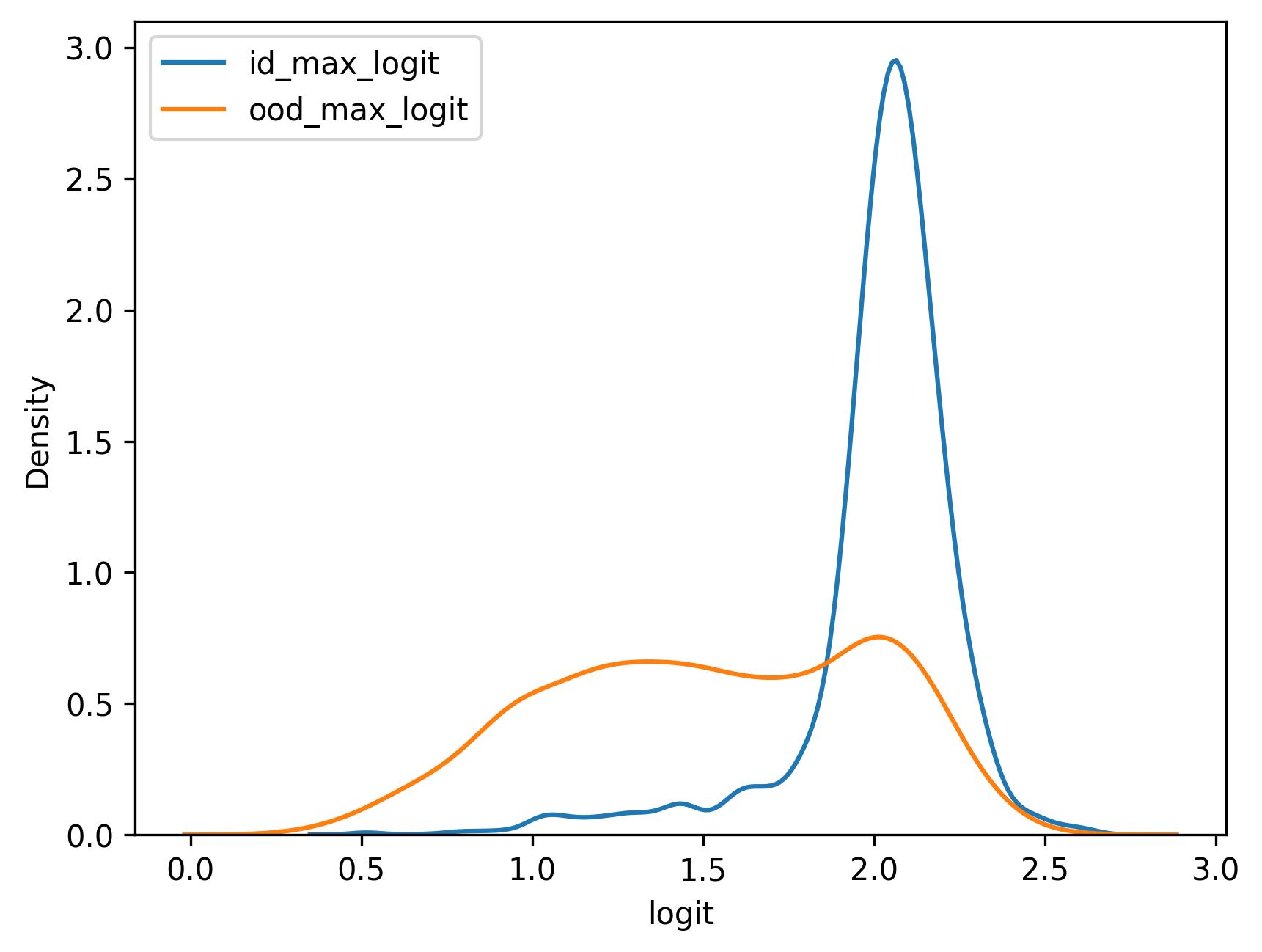}}
    \subfigure[ALS 5]{\includegraphics[width=0.16\textwidth]{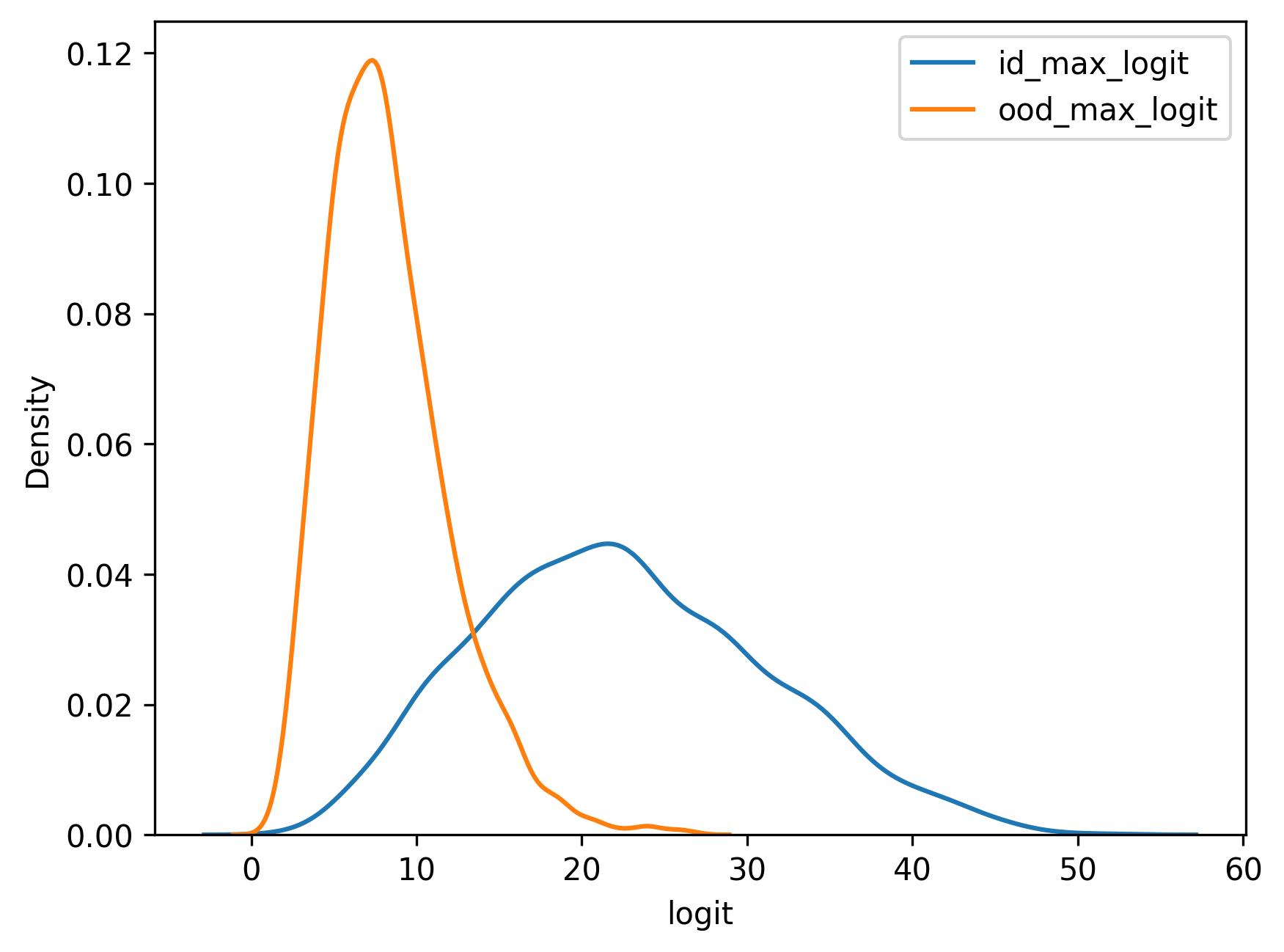}}
    \subfigure[ALS 10]{\includegraphics[width=0.16\textwidth]{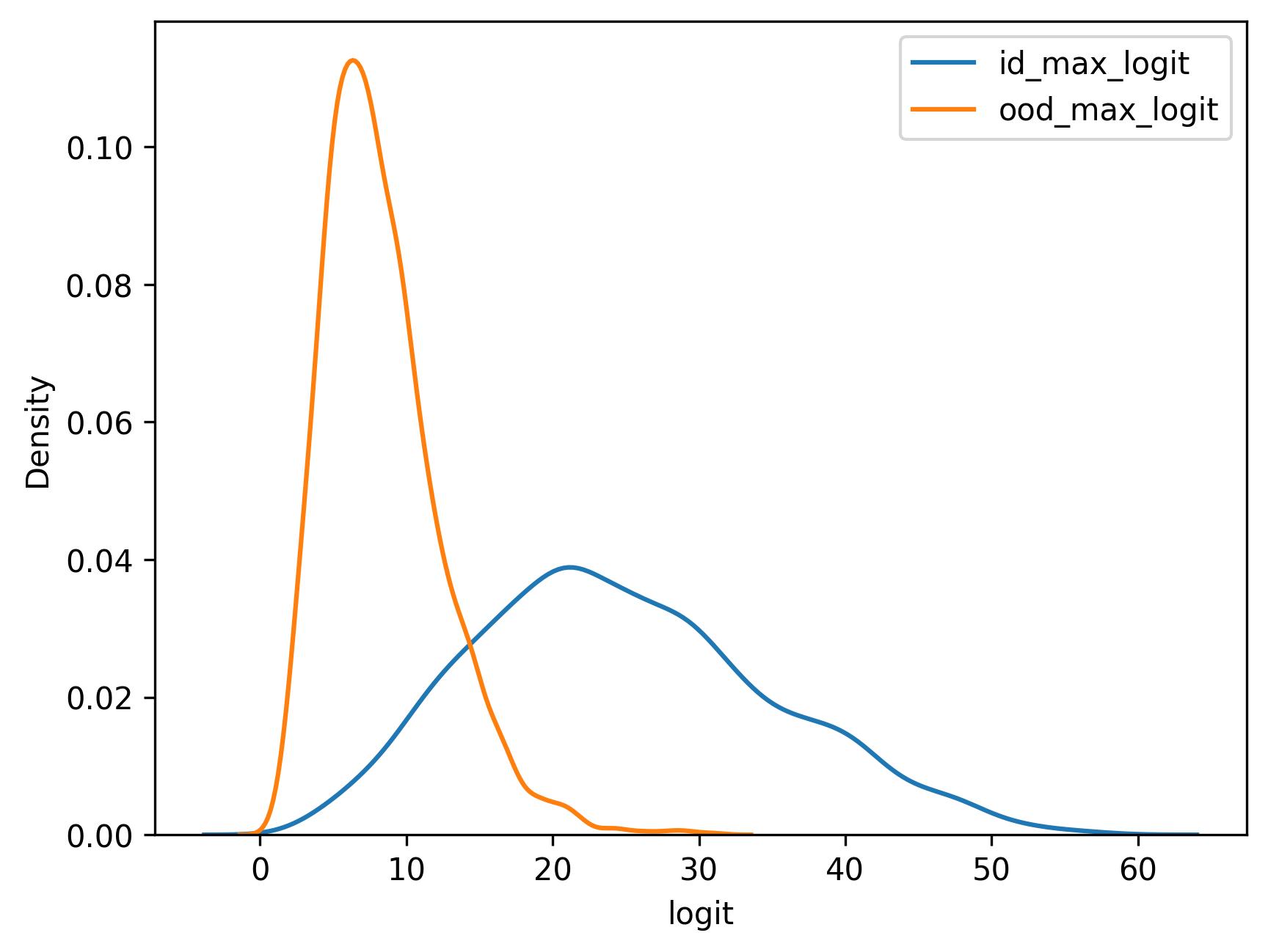}}
    \subfigure[ALS 20]{\includegraphics[width=0.16\textwidth]{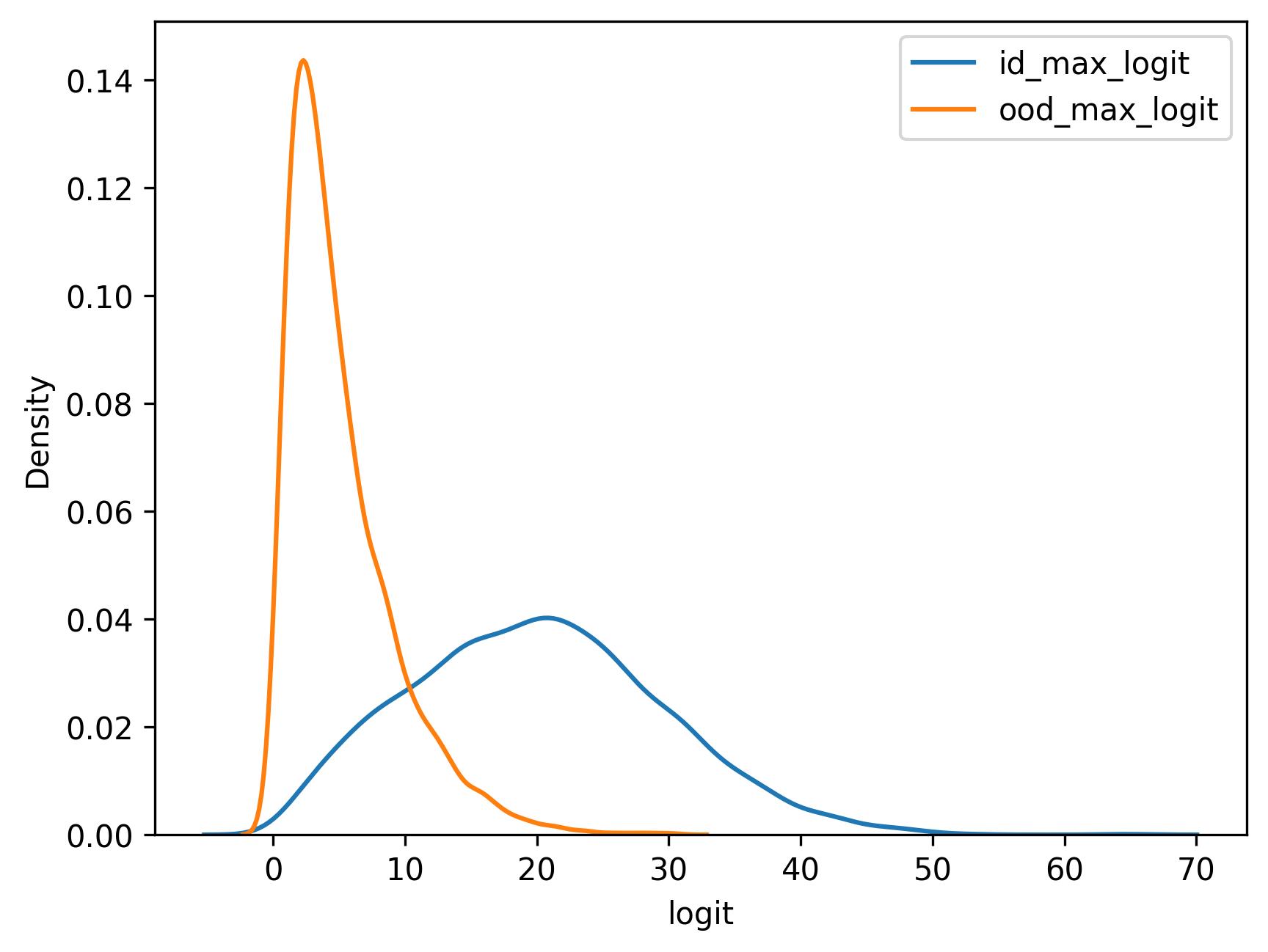}}
\caption{First and second rows visualize the known and all test samples (including known and unknown) via t-SNE. The third and fourth rows illustrate the density distribution of unknown score with maximal probability and logit as the score function. Each column shows a method trained in CIFAR10. Zoom in to see the scales.}
\label{fig1}
\end{figure*}

In this paper, we find that label smoothing shapes the distribution spaces for predicted probability and logit (before Softmax). As shown in Figure \ref{fig1}, the scale of maximal probability and maximal logit become smaller, by which unknown and known samples have more overlap and the OOD detection performance thus degrades. A similar observation is discussed \cite{muller2019does}, but in the context of knowledge distillation. To understand and overcome this issue deeper, we revisit label smoothing from a disentangled perspective. When training a model with label smoothing, a true class is assigned a predefined target slightly smaller than one, and another \emph{same} tiny target, slightly larger than zero, is assigned for all non-true classes. This soft target for the true class results in the smaller scale of maximal logit and probability. In contrast, the tiny equal soft target for all non-true classes is the core to contribute classification, by which the model is pushed to learn features useful only for the true class and not useful to distinguish the input into one of the non-true classes. In other words, the learned features are not shared \emph{essentially} and therefore are more discriminative \cite{perera2019deep}.

With this understanding, we further adapt label smoothing to facilitate OOD detection. For true classes, we argue that the target should neither be limited to a smaller value than one nor fixed for all training samples considering the uncertainty \cite{malinin2018predictive, sensoy2018evidential, ulmer2023prior}. Simultaneously, the non-true classes should be designed with same target to have non-shared feature in essential. However, the desired learning target for the true class has no ground truth because the uncertainty is generally unknown. Rather than explicitly making a learning target, we propose a new regularization to push the non-true class to have a same target probability through minimizing their standard deviation. In terms of the true class, the cross-entropy loss is borrowed to learn the uncertainty from empirical data. In this setting, the benefit of label smoothing is retained and the disadvantage is mitigated.
To conclude, our contribution is as follows.
\begin{itemize}
    \item We find that the limited and predefined learning target for the true class in label smoothing degrades the performance of OOD detection.
    \item We propose a novel method to address the issue by regularizing the learned probabilities, called adaptive label smoothing (ALS), where the maximal probabilities is neither limited nor fixed and the non-true classes are pushed to have same probabilities. 
    \item Comprehensive experimental results across six datasets with different score functions show that the proposed ALS contributes to OOD detection with a clear margin, better performance for classifying known and discerning unknown from known.
\end{itemize}

\section{Related Work}

\subsection{Regularization}
Large-scale models have contributed in the last decades. One of the potential issues is overfitting due to many more learnable parameters than training data. Regularization can be used to mitigate this problem \cite{srivastava2014dropout, gelman2021most}. Dropout \cite{srivastava2014dropout} randomly blocks some units and their connections, which allows the model not to rely on some information. The weight decay \cite{krogh1991simple} aims to reduce the irrelevant weight to improve generalization. In the context of multi-class classification, label smoothing \cite{szegedy2016rethinking} creates a soft learning target where non-true class has a tiny target and a true class has a target slightly smaller than one. With this target, models tend to be less confident. Although label smoothing contributes to a compact intra-class feature space, it results in a smaller scale of logits before a Softmax layer and degrades some downstream tasks such as knowledge distillation \cite{muller2019does}. Empirical results suggest that label smoothing tends to harm OOD detection \cite{vaze2022open}. In this paper, our objective is to explain the potential reasons and then overcome the challenge of improving OOD detection.

\subsection{OOD Detection}
OOD essentially consists of two tasks, recognizing unknown from known and classifying known. The second task can be achieved using the softmax layer with the predicted class by choosing the classes with the maximal probabilities. Therefore, a fundamental topic in OOD detection is to design a score function for the former task. In the core, the score function is to measure the uncertainty \cite{malinin2018predictive, sensoy2018evidential, ulmer2023prior}. A baseline score function is the maximal probability \cite{hendrycks2016baseline} assuming that the unknown has a smaller maximal probability. Similarly, the Shannon entropy \cite{liu2023gen} is another uncertainty measure. To be more sensitive for probabilities lower than one, generalized entropy (GEN) is used in \cite{liu2023gen}. Since the distribution of predicted probability may depend on classes, the class-conditional probability is introduced in KL-matching \cite{hendrycks2022scaling} where the score function is the minimal KL divergence in the probability space between the test sample and the other center of the class.

Except for probability space, logit as a non-relative measure shows its superiority. The straightforward is the maximal logit \cite{hendrycks2022scaling, vaze2022open} where known has a larger logit than unknown. As an uncertainty measure, the energy defined in the logits is used in \cite{liu2020energy}. Besides, feature space is also used to design score functions, such as Mahalanobis distance \cite{lee2018simple}. Considering the heterogeneous contribution to the probability, logit, and probability level \cite{zhang2023decoupling}, combining them together provides a chance such as ViM \cite{wang2022vim}. Furthermore, other levels suggests the hints between known and unknowns, such as gradient \cite{huang2021importance}. Our method aims to train a better model with only known data and is therefore compatible with diverse score functions.

A core challenge for OOD detection is the unavailability of unknown samples when training. Consequently, a branch is to probe how unknown can be recognized and then to design the strategy to improve it. The familiarity hypothesis \cite{dietterich2022familiarity} declares that the trained classifier learns the feature for the known classes and then the learned feature is the core to separate unknown from known. To be more specific, those samples with similar learned feature is considered as known and verse vice, without additional feature from unknown samples. In this direction, making use of the known dataset is embraced. Empirically, a model with better performance in known data tends to be more decent in classifying unknown from known \cite{vaze2022open}. Furthermore, achieving the desired feature space, compact for intra-class and disperse for inter-class, is also convincing \cite{ming2023how, liu2022orientational, meng2023known}. In this paper, we consider regularization to facilitate OOD detection.

\section{Method}

\subsection{Formalization and Notation}
For a multi-class classification, let $X$ and $Y$ denote input and label distribution spaces. A model is first optimized on a training dataset $\mathcal{D}_{tr}$ sampled from a joint distribution $X_{tr} \times Y_{tr}$ and then evaluated on a test dataset $\mathcal{D}_{te}$ sampled from a joint distribution $X_{te} \times Y_{te}$. Under the identical and independent distribution (i.i.d.) assumption, $X_{tr} = X_{te}$ and $Y_{tr} = Y_{te}$. The task is to classify a new input $\bm{x}$ in the corresponding label $\bm{y}$ in one-hot manner. Cross-entropy is a widely used loss function to optimize models and is formulated as:
\begin{equation}
     H(\bm{y}, \bm{p}) = -\sum_{i=1}^N y_i \log p_i,
\end{equation}
where $\bm{p}$ is the predicted probability and $p_i$ is the predicted probability for class $i$. $N$, the total number of known classes, is the cardinality of $Y_{tr}$. $y_i$ is 1 for the correct class and 0 for the rest. The cross-entropy loss forces a model to have probability one for the true class and to have probability zero for the non-true class.

However, the i.i.d. assumption is often conflicted in many applications. In contrast, OOD detection allows $Y_{te} = Y_{tr} + U$ where $U$ is not empty. In general, $U$ is called a super-unknown class. In this scenario, two tasks exist: distinguishing unknown classes from known classes and classifying the known classes into the corresponding ones. To achieve the tasks, two score and decision functions are used \cite{wang2022openauc}. For the second known classification task, the common score function $s_k$ is predicted probabilities and the decision function $d_k$ is to chose the class with the maximal predicted probability, such that $\hat{y} = \mathop{\arg\max}_i p_i$. Recognizing unknown is as follows:
\begin{equation}
    \bm{x} \in
        \begin{cases}
            Y_{tr} & \text{if } (d_{uk} \circ s_{uk})(\bm{x}) \le \tau \, ; \\
            U & \text{otherwise} \, , \\
        \end{cases}
\label{unknown_function}
\end{equation}
where $\circ$ denotes a composition function and $\tau$ is a given threshold. $d_{uk}$ and $s_{uk}$ suggest unknown decision and score functions.


\subsection{OOD Detection Risks}
With the unknown score and the decision function in Equation \ref{unknown_function} such as the maximal probability \cite{hendrycks2016baseline}, two risks exist \cite{scheirer2011meta}. First, known samples with smaller scores partly due to the uncertainty of the data \cite{ulmer2023prior} are risky to be classified as unknown. For example, an occluded object may have less evidence, as well as hard examples. In this case, data augmentation is an effective strategy \cite{xu2023comprehensive}. As shown in \cite{vaze2022open}, strong data augmentation with longer training benefits OOD detection.

The second risk is that unknown samples that have high scores are risky to be classified as known. The primary reason is that unknown samples have the same characteristics as known samples, such as learned features or activation \cite{dietterich2022familiarity}. To mitigate this issue, a paradigm is using unknown data \cite{liu2020energy, dhamija2018reducing} or synthetic unknown data \cite{moon2022difficulty, jiang2023openmix, xu2023contrastive, wang2024learning, neal2018open}, to learn more discriminative features. Another strategy is to learn non-shared feature for each known class. For example, membership loss \cite{perera2019deep} was proposed to penalize the non-true class with high activation. Similarly, compact intra-class and disperse inter-class feature spaces are assumed to be desired \cite{ming2023how, liu2022orientational, meng2023known}. In this paper, our aim is to facilitate this risk by learning features that are only useful to classify the input into the true class from all classes, yet not useful to distinguish the input into one non-true class from another non-true class.

\subsection{Revisiting Label Smoothing}

The vanilla cross-entropy loss uses one hot label as the learning target and this hard target may trigger overfitting and degrade the generalization, label smoothing (LS) was introduced in \cite{szegedy2016rethinking} to mitigate this issue, by using a soft target $\bm{\tilde{y}}$:
\begin{equation}
    \bm{\tilde{y}} = \underbrace{(1-\alpha) \bm{y}}_{\text{true class}} + \underbrace{\frac{\alpha}{N-1} (1 - \bm{y})}_{\text{non-true class}}
\label{label_smoothing_equation}
\end{equation}
where $\alpha \in [0, 1]$ is a hyper-parameter. Empirically, label smoothing contributes to recognition performance for known classes \cite{szegedy2016rethinking} but may alter downstream tasks, such as knowledge distillation \cite{muller2019does} and OOD detection \cite{lee2020soft, vaze2022open}. To probe this phenomenon, we rethink label smoothing from a disentangled perspective.

First, label smoothing pushes non-true classes to have the same target. To achieve it, models try to learn those features only useful to discern the true class from all other known classes yet not for any non-true known classes. It leads to better learning feature spaces for those known classes. As shown in Figure \ref{fig1}, t-SNE \cite{van2008tSNE} suggests that label smoothing produces a feature space, compact intra-class and larger inter-class disparity. Similar observation is reported in \cite{muller2019does}.

Second, label smoothing decreases the target for the true class, which will further push the maximal probability and the maximum logit to smaller \cite{muller2019does}. As shown in Figure \ref{fig1}, the mean probability of the maximum converges to $1-\alpha$ and the maximum logit also shrinks much, where the unknown may have more overlap with the known. Consequently, the performance to distinguish the unknown is impaired. In addition, we argue that this hard target is not reasonable when classifying samples in diverse difficulties. In contrast, it is more convincing to assign a higher target for the easy samples and a lower one for the hard ones.

\subsection{Adaptive Label Smoothing}
\label{als}

Based on the above analysis, we propose adaptive label smoothing to adapt label smoothing (ALS) to cater to OOD detection. The adaptive reflects in two ways. First, the maximal probability should be neither limited, to have large scale probability and logit for discerning unknown from known, nor a fixed values, to consider the diversity of classification difficulties. Second, the classes with non-maximal probability should be pushed to have the same level of probability for a given sample, in spite of diversity for different samples. Furthermore, we do not make a target probability distribution, since it should be sample-specific and is generally not known. We instead regularize the predicted probabilities. Conceptually, our adaptive label smoothing consists of two parts accordingly and is formulated as:
\begin{equation}
    \mathcal{L}_{ALS} = \mathcal{L}_{MPC} + \lambda \cdot \mathcal{L}_{NMPC},
\label{adaptive_ls}
\end{equation}
where $\mathcal{L}_{MP}$ and $\mathcal{L}_{NMP}$ suggest the loss for maximal probability class (MPC) and non-maximal probability class (NMPC), respectively. A hyper-parameter $\lambda$ balances the two parts. We directly borrow cross-entropy loss for the first part to push true class as MPC, such that $\mathcal{L}_{MPC} = H(\bm{y}, \bm{p})$. For the second part, we aim to push the NMPC to have same probabilities, regardless of the value of maximal probability:
\begin{equation}
    \mathcal{L}_{NMPC} = \sqrt{\frac{1}{N-1} \sum_{i=1,i \neq k}^N(p_i - \bar{p})^2},
\end{equation}
where $\bar{p} = \frac{1}{N-1} \sum_{i=1,i \neq k}^N p_i$ is the average predicted probability for NMPC and $k=\mathop{\arg \max}_i p_i$. In short, the probabilities for the NMPC are pushed to have standard deviation zero.

$L_{NMPC}$ elusively embraces that the MPC is the true class and may be thus risky when it is not the case. Several strategies can be used. First, $L_{NMPC}$ is only adopted when the assumption is the true case. Second, $\lambda$ is gradually enlarged to wait the true classes having the maximal probabilities. Experimental results suggest that the strategies make the training stable and the first one is superior to the second.

\textbf{Discussion.} We argue that our proposed method facilitates the second OOD detection risk. Intuitively, radically pushing the true class with probability one and non-true class with zero is difficult because of the natural uncertainty in data. In general, the true class and non-true classes may share some features and therefore, the true class has a probability smaller than one and the closer non-true classes have a probability bigger than zero. Instead, we regularize all of non-true classes with the same probabilities, allowing the existence of shared features yet with no contribution to classify the input into one of the non-true classes. In this case, shared features are essentially equivalent to non-shared features. From the Figures \ref{fig1}, ALS contributes both known and unknown classifications, with a compact intra-class and disparate inter-class feature space. Although LS also pushes the similar feature space, it costs at reduced scales of maximal probability and logits, with which unknown may have the similar scores as known and thus the OOD detection degrades. In addition, we noticed that an existing paper \cite{krothapalli2021one} used the same name as ours, in which the area ratio between the cropped and the original images is used to make the soft target and the core is thus totally different from ours.

\section{Experiment}

\begin{table*}[!ht]
    \centering
    \begin{tabular}{|l|ll|ll|ll|ll|ll|ll|}
        \hline
        & \multicolumn{2}{|c|}{MNIST} & \multicolumn{2}{|c|}{SVHN} & \multicolumn{2}{|c|}{CIFAR10} & \multicolumn{2}{|c|}{CIFAR100} & \multicolumn{2}{|c|}{TinyIN} & \multicolumn{2}{|c|}{average} \\
        \cline{2-13}
        & FPR & OSCR & FPR & OSCR & FPR & OSCR & FPR & OSCR & FPR & OSCR & FPR & OSCR \\ 
        \hline
        \multicolumn{13}{|l|}{score function: max prob \cite{hendrycks2016baseline}} \\
        baseline & \textbf{4.18} & \textbf{98.57} & 29.09 & 94.04 & 54.72 & 89.39 & 28.40 & 94.57 & 77.63 & 71.70 & 38.80 & 89.65 \\ 
        LS (0.1) & 9.07 & 93.46 & \textbf{20.39} & 92.72 & 54.93 & 82.17 & 32.70 & 86.98 & 77.33 & 72.02 & 38.88 & 85.47 \\ 
        LS (0.3) & 10.60 & 92.78 & 21.27 & 91.41 & 54.73 & 79.91 & 33.12 & 84.22 & 76.67 & 71.01 & 39.28 & 83.87 \\ 
        ALS (ours) & 4.58 & 98.56 & 24.86 & \textbf{94.55} & \textbf{51.65} & \textbf{89.82} & \textbf{23.06} & \textbf{95.07} & \textbf{75.84} & \textbf{72.62} & \textbf{36.00} & \textbf{90.12} \\ 
        \hline
        \multicolumn{13}{|l|}{score function: Entropy} \\
        baseline & \textbf{3.99} & \textbf{98.63} & 27.48 & 94.26 & 53.11 & 89.59 & 27.40 & 94.74 & 76.42 & 72.30 & 37.68 & 89.90 \\ 
        LS (0.1) & 9.27 & 93.39 & \textbf{19.94} & 92.62 & 53.05 & 81.90 & 32.10 & 86.86 & 75.44 & 72.23 & 37.96 & 85.40 \\ 
        LS (0.3) & 10.71 & 92.72 & 21.14 & 91.25 & 53.42 & 79.64 & 33.68 & 84.09 & 77.48 & 70.54 & 39.29 & 83.65 \\ 
        ALS (ours) & 4.50 & 98.60 & 24.23 & \textbf{94.69} & \textbf{51.17} & \textbf{90.28} & \textbf{22.66} & \textbf{95.24} & \textbf{73.58} & \textbf{73.15} & \textbf{35.23} & \textbf{90.39} \\ 
        \hline
        \multicolumn{13}{|l|}{score function: GEN \cite{liu2023gen}} \\
        baseline & \textbf{2.73} & \textbf{98.94} & 19.60 & 94.86 & 41.94 & 90.93 & \textbf{15.72} & \textbf{95.80} & 74.50 & 72.17 & 30.90 & 90.54 \\ 
        LS (0.1) & 9.61 & 93.16 & 20.53 & 92.13 & 53.58 & 81.08 & 32.52 & 86.51 & 75.44 & 71.22 & 38.34 & 84.82 \\ 
        LS (0.3) & 10.85 & 92.64 & 21.10 & 90.98 & 53.61 & 79.25 & 33.76 & 83.95 & 78.27 & 69.58 & 39.52 & 83.28 \\ 
        ALS (ours) & 3.32 & 98.85 & \textbf{18.22} & \textbf{95.09} & \textbf{39.61} & \textbf{91.31} & 17.66 & 95.75 & \textbf{74.33} & \textbf{73.20} & \textbf{30.63} & \textbf{90.84} \\ 
        \hline
        \multicolumn{13}{|l|}{score function: max logit \cite{hendrycks2022scaling, vaze2022open}} \\
        baseline & \textbf{2.39} & \textbf{98.99} & 19.06 & 94.87 & 41.12 & 91.03 & \textbf{11.46} & \textbf{96.14} & 75.74 & 71.64 & 29.95 & 90.53 \\ 
        LS (0.1) & 9.71 & 93.06 & 20.30 & 92.00 & 53.63 & 80.96 & 32.72 & 86.27 & 77.06 & 71.66 & 38.68 & 84.79 \\ 
        LS (0.3) & 10.81 & 92.66 & 21.28 & 91.09 & 54.16 & 79.45 & 33.52 & 83.97 & 76.80 & 70.67 & 39.31 & 83.57 \\ 
        ALS (ours) & 3.12 & 98.89 & \textbf{17.46} & \textbf{95.10} & \textbf{37.98} & \textbf{91.46} & 14.02 & 95.97 & \textbf{73.92} & \textbf{72.63} & \textbf{29.30} & \textbf{90.81} \\ 
        \hline
        \multicolumn{13}{|l|}{score function: energy \cite{liu2020energy}} \\
        baseline & \textbf{2.38} & \textbf{99.00} & 19.00 & 94.86 & 41.04 & 91.00 & \textbf{11.28} & \textbf{96.15} & 75.46 & 71.32 & 29.83 & 90.47 \\ 
        LS (0.1) & 9.83 & 93.02 & 20.75 & 91.74 & 53.52 & 80.46 & 32.86 & 86.12 & 76.10 & 70.87 & 38.61 & 84.44 \\ 
        LS (0.3) & 10.96 & 92.60 & 21.27 & 90.81 & 53.90 & 78.99 & 33.86 & 83.79 & 78.61 & 69.41 & 39.72 & 83.12 \\ 
        ALS (ours) & 3.12 & 98.89 & \textbf{17.34} & \textbf{95.10} & \textbf{37.75} & \textbf{91.45} & 13.88 & 95.99 & \textbf{74.69} & \textbf{72.37} & \textbf{29.36} & \textbf{90.76} \\ 
        \hline
        \multicolumn{13}{|l|}{score function: React \cite{sun2021react}} \\
        baseline & \textbf{2.38} & \textbf{99.00} & 18.99 & 94.86 & 41.04 & 91.00 & 11.28 & \textbf{96.15} & 75.46 & 71.32 & 29.83 & 90.47 \\ 
        LS (0.1) & 10.16 & 92.95 & 20.69 & 91.80 & 53.46 & 80.66 & 33.26 & 86.35 & 76.10 & 70.87 & 38.73 & 84.53 \\ 
        LS (0.3) & 11.05 & 92.52 & 21.21 & 90.84 & 54.18 & 79.19 & 36.76 & 83.71 & 78.61 & 69.41 & 40.36 & 83.13 \\ 
        ALS (ours) & 3.12 & 98.89 & \textbf{17.34} & \textbf{95.10} & \textbf{37.74} & \textbf{91.45} & 13.88 & 95.99 & \textbf{74.69} & \textbf{72.38} & \textbf{29.35} & \textbf{90.76} \\ 
        \hline
        \multicolumn{13}{|l|}{score function: grad norm \cite{huang2021importance}} \\
        baseline & 21.13 & 94.39 & 36.33 & 88.16 & 54.45 & 85.44 & 21.78 & 94.50 & 83.80 & 59.58 & 43.50 & 84.41 \\ 
        LS (0.1) & 18.68 & 91.03 & 20.65 & \textbf{94.59} & \textbf{45.32} & \textbf{89.85} & 23.98 & \textbf{94.96} & \textbf{83.69} & 57.68 & \textbf{38.46} & 85.62 \\ 
        LS (0.3) & 20.45 & 89.06 & \textbf{19.48} & 94.44 & 46.27 & 89.64 & 25.22 & 94.39 & 84.01 & \textbf{60.36} & 39.09 & 85.58 \\ 
        ALS (ours) & \textbf{14.32} & \textbf{96.52} & 29.16 & 91.34 & 52.09 & 86.39 & \textbf{19.66} & 94.93 & 84.24 & 59.82 & 39.89 & \textbf{85.80} \\ 
        \hline
        \multicolumn{13}{|l|}{score function: ViM \cite{wang2022vim}} \\
        baseline & \textbf{2.37} & \textbf{99.00} & 18.87 & 94.87 & 41.60 & 90.99 & \textbf{11.24} & \textbf{96.16} & 74.56 & 71.55 & 29.73 & 90.51 \\ 
        LS (0.1) & 9.56 & 93.38 & 20.67 & 91.87 & 54.99 & 79.45 & 33.68 & 85.70 & 76.16 & 71.04 & 39.01 & 84.29 \\ 
        LS (0.3) & 10.57 & 93.36 & 21.48 & 90.94 & 56.25 & 77.26 & 34.90 & 83.12 & 80.60 & 69.16 & 40.76 & 82.77 \\ 
        ALS (ours) & 3.12 & 98.89 & \textbf{17.19} & \textbf{95.11} & \textbf{37.74} & \textbf{91.44} & 13.76 & 95.99 & \textbf{74.43} & \textbf{72.59} & \textbf{29.25} & \textbf{90.80} \\ 
        \hline
    \end{tabular}
\caption{Performance across five datasets and their averages. LS and ALS suggest label smoothing and our proposed adaptive label smoothing. The value for LS denotes the hyper-parameter $\alpha$. The bold face shows the best performance in each case. We report our proposed ALS with hyper-parameter $\lambda=5$ that may not be the optimal.}
\label{table2}
\end{table*}

\begin{table*}[!ht]
    \centering
    \begin{tabular}{|l|l|l|l|l|l|l|l|l|l|}
        \hline
        & \multirow{2}{*}{baseline} & \multicolumn{3}{|c|}{label smoothing (LS)} & \multicolumn{4}{c|}{adaptive label smoothing (ALS)} \\
        \cline{3-9}
        & & $\alpha = 0.1$ & $\alpha = 0.2$ & $\alpha = 0.3$ & $\lambda = 1$ & $\lambda = 5$ & $\lambda = 10$ & $\lambda = 20$ \\
        \hline
        MNIST & 99.60 & \textbf{\underline{99.63}} & \textbf{99.62} & \textbf{\underline{99.63}} & \textbf{99.61} & 99.57 & 99.58 & 99.57 \\ 
        SVHN & 97.45 & \textbf{97.88} & \textbf{\underline{98.01}} & \textbf{97.98} & 97.43 & \textbf{97.52} & \textbf{97.71} & \textbf{97.83} \\ 
        CIFAR10 & 96.55 & \textbf{\underline{96.86}} & \textbf{96.83} & \textbf{96.85} & \textbf{96.58} & \textbf{96.67} & \textbf{96.66} & 96.32 \\ 
        CIFAR100 & 97.80 & \textbf{\underline{98.06}} & \textbf{98.20} & \textbf{98.22} & \textbf{98.00} & \textbf{97.92} & \textbf{97.86} & \textbf{97.92} \\ 
        TinyIN & 82.74 & 82.56 & 82.26 & 81.66 & 82.74 &\textbf{83.46} & \textbf{\underline{83.64}} & \textbf{82.94} \\ 
        \hline
        average & 94.83 & \textbf{95.00} & \textbf{94.98} & \textbf{94.87} & \textbf{94.87} & \textbf{95.03} & \textbf{\underline{95.09}} & \textbf{94.92} \\ 
        \hline
    \end{tabular}
\caption{Known test accuracy. The bold face suggests the superior accuracy than the baseline and the underlined is the best.}
\label{table1}
\end{table*}

\begin{table}[!ht]
    \centering
    \begin{tabular}{|l|lll|lll|}
        \hline
        & m@1 & m@2 & m@3 & m@1 & m@2 & m@3 \\ \hline
        & \multicolumn{3}{|c|}{max prob } & \multicolumn{3}{|c|}{energy} \\
        baseline & 92.09 & 38.80 & 89.65 & 93.59 & 29.83 & 90.47 \\ 
        ALS 1 & 92.24 & 38.23 & 89.81 & 93.71 & 29.45 & 90.59 \\ 
        ALS 5 & 92.49 & 36.00 & 90.12 & \textbf{93.73} & \textbf{29.36} & \textbf{90.76} \\ 
        ALS 10 & \textbf{92.61} & 34.79 & \textbf{90.27} & 93.61 & 30.39 & 90.72 \\ 
        ALS 20 & 92.60 & \textbf{33.98} & 90.11 & 93.24 & 32.43 & 90.32 \\ 
        \hline
        & \multicolumn{3}{|c|}{entropy} & \multicolumn{3}{|c|}{React} \\
        baseline & 92.47 & 37.68 & 89.90 & 93.60 & 29.83 & 90.47 \\ 
        ALS 1 & 92.62 & 37.00 & 90.05 & 93.71 & 29.44 & 90.59 \\ 
        ALS 5 & 92.88 & 35.23 & 90.39 & \textbf{93.73} & \textbf{29.35} & \textbf{90.76} \\ 
        ALS 10 & \textbf{93.05} & 33.81 & \textbf{90.57} & 93.61 & 30.39 & 90.72 \\ 
        ALS 20 & 93.00 & \textbf{32.93} & 90.39 & 93.24 & 32.42 & 90.32 \\ 
        \hline
        & \multicolumn{3}{|c|}{GEN } & \multicolumn{3}{|c|}{grad norm} \\
        baseline & 93.55 & 30.90 & 90.54 & 87.44 & 43.50 & 84.41 \\ 
        ALS 1 & 93.66 & 30.73 & 90.65 & 87.98 & 42.67 & 84.91 \\ 
        ALS 5 & \textbf{93.70} & \textbf{30.63} & \textbf{90.84} & 88.77 & 39.89 & 85.80 \\ 
        ALS 10 & 93.65 & 31.02 & \textbf{90.84} & 89.40 & 38.30 & 86.49 \\ 
        ALS 20 & 93.34 & 32.47 & 90.47 & \textbf{89.80} & \textbf{36.44} & \textbf{86.81} \\ 
        \hline
        & \multicolumn{3}{|c|}{max logit} & \multicolumn{3}{|c|}{ViM} \\
        baseline & 93.62 & 29.95 & 90.53 & 93.64 & 29.73 & 90.51 \\ 
        ALS 1 & \textbf{93.74} & 29.45 & 90.66 & 93.76 & 29.44 & 90.64 \\ 
        ALS 5 & \textbf{93.74} & \textbf{29.30} & \textbf{90.81} & \textbf{93.77} & \textbf{29.25} & \textbf{90.80} \\ 
        ALS 10 & 93.61 & 30.51 & 90.77 & 93.65 & 30.38 & 90.76 \\ 
        ALS 20 & 93.24 & 32.44 & 90.35 & 93.28 & 32.48 & 90.36 \\ 
        \hline
    \end{tabular}
\caption{Impact of hyper-parameter in adaptive label smoothing. The average AUROC (m@1) FPR (m@2), and OSCR (m@3) on five datasets are reported. Bold face denotes the best for each case.}
\label{table3}
\end{table}

\subsection{Implementation}

\textbf{Dataset and optimization}. Our implementation is the same as in a previous paper \cite{vaze2022open}. MNIST \cite{mnist}, SVHN \cite{svhn}, CIFAR10 and CIFAR100 \cite{cifar10}, TinyIN (Tiny-ImageNet) \cite{tinyimagenet}, are used as benchmark datasets. For MNIST, SVHN, and CIFAR10, six classes are randomly selected as known, and the other four classes as unknown. For CIFAR100, 4 non-shared classes are chosen from CIFAR10 as known and 10 unknown classes are selected from CIFAR100. With TinyIN, 20 known classes are taken as known and remained 180 classes are taken as unknown. Images are in resized as 32*32 for all datasets, except for TinyIN with 64*64. The learning rate is set as 0.01 for TinyIN and 0.1 for other datasets, with cosine annealing schedule. All datasets are trained with VGG32 \cite{vgg}, 600 epochs, batch size 128, and rand-augment \cite{cubuk2020randaugment}. Every dataset is trained with random five splits and the average performance is reported.

\textbf{OOD score function}. Our method can be integrated with other OOD score functions. The following are adopted: max probability (max prob) \cite{hendrycks2016baseline}, Shannon entropy (entropy), GEN \cite{liu2023gen}, max logit \cite{hendrycks2022scaling, vaze2022open}, energy \cite{liu2020energy}, React \cite{sun2021react}, grad norm \cite{huang2021importance}, ViM \cite{wang2022vim}. Those score functions achieve current state-of-the-art performance in OOD detection. For GEN, all probabilities are used and the gamma 0.1 is taken as the original paper suggested \cite{liu2023gen}.

\textbf{OOD evaluation metric}. Accuracy, AUROC, FPR with TPR at 0.95, OSCR \cite{dhamija2018reducing} are adopted to evaluate OOD methods. Accuracy (acc) is computed when the test data only include known classes. In general, AUROC is used to evaluate binary classification, known or unknown. Although it is threshold-free, it does not evaluate OOD detection completely. OSCR \cite{dhamija2018reducing} is introduced to evaluate two tasks simultaneously, correct classification among known classes, and unknown from known. The original OSCR is not threshold-free and can be adapted as AUROC to get a single value \cite{wang2022openauc}. FPR (false positive rate) when TPR (true positive rate) is 95 percent is denoted as FPR. The bigger acc, AUROC, and OSCR, the better OOD method. In contrast, a smaller FPR leads to a superior OOD method.

\subsection{Main Result}
\textbf{Quantitative comparison}. As shown in Table \ref{table1}, adaptive label smoothing results in better accuracy of known data, with suitable hyper-parameter. An observation is that the label smoothing in TinyIN is inferior to the baseline, whereas our method is still superior. For our method, the known accuracy tends to grow and reduce when $\lambda$ grows gradually. The performance of OOD detection is given in Table \ref{table2}. As suggested in a previous article \cite{vaze2022open}, label smoothing has slightly better performance in TinyIN and tends to be harmful in other datasets with a clear margin. For example, the average OSCR degrades to 85.47 from 89.65 with max prob and becomes 84.79 from 90.53 with max logit. In contrast, our method secures the best FPR and OSCR. An exception is in the grad norm, where label smoothing achieves the lowest FPR. Our method outperforms the baseline very well, no matter which score functions. Based on the performance, we argue that our proposed adaptive label smoothing benefits both known and unknown classifications. Our method is superior to the state-of-the-art in terms of the score functions across the five datasets.

\textbf{Visualization}. As shown in Figure \ref{fig1}, label smoothing and our method result in a better feature space, such as intra-class compactness and inter-class disparity. Adaptive label smoothing does not shrink the scale of max probability and logits as the label smoothing does. Besides, our proposed adaptive label smoothing contributes to a larger scale of maximal logits. We also observe that, although feature space are more compact for intra-class end more disparate for known samples, the OOD detection is not better. We conjecture that mode collapse may exist, where unknown features are more similar to the known ones. In other words, the feature of OOD samples also locates in the compact feature spaces of known sample. This observation supports that OOD detection should adopt the joint distribution of input and label space not just the marginal distribution such as input space. Furthermore, the results suggest that a good closed set classifier may not always contribute to the OOD detection as declared in \cite{vaze2022open}.

\subsection{Analysis on adaptive label smoothing}

\begin{table}[!ht]
    \centering
    \begin{tabular}{|l|lll|lll|}
    \hline
        variant & \multicolumn{3}{c|}{CIFAR10} & \multicolumn{3}{c|}{TinyIN} \\ \hline
        ~ & m@1 & m@2 & m@3 & m@1 & m@2 & m@3 \\ 
        \hline
        & \multicolumn{6}{l|}{max prob} \\
        all 0 & 91.38 & 50.77 & 89.93 & 80.14 & 78.46 & 72.01 \\ 
        all 20 & 91.52 & 49.84 & 89.91 & \textbf{80.55} & 76.76 & 72.54 \\ 
        all 50 & \textbf{91.60} & \textbf{49.43} & \textbf{90.12} & 80.07 & 77.97 & 72.12 \\ 
        all 100 & 91.43 & 51.86 & 89.98 & 80.00 & 79.19 & 72.09 \\ 
        only corr & 91.31 & 51.65 & 89.82 & 80.33 & \textbf{75.84} & \textbf{72.62} \\ 
        \hline
        & \multicolumn{6}{l|}{entropy} \\
        all 0 & 91.92 & 49.94 & 90.39 & 81.13 & \textbf{73.18} & 72.54 \\ 
        all 20 & \textbf{92.07} & 49.24 & 90.40 & \textbf{81.35} & 74.43 & 73.01 \\ 
        all 50 & \textbf{92.07} & \textbf{48.25} & \textbf{90.57} & 80.98 & 74.63 & 72.58 \\ 
        all 100 & 91.95 & 50.73 & 90.42 & 80.97 & 74.88 & 72.59 \\ 
        only corr & 91.82 & 51.17 & 90.28 & 81.23 & 73.58 & \textbf{73.15} \\ 
        \hline
        & \multicolumn{6}{l|}{GEN} \\
        all 0 & 93.03 & 40.41 & 91.15 & 82.02 & 73.99 & 72.43 \\ 
        all 20 & 93.29 & 39.70 & 91.21 & \textbf{82.30} & 73.67 & 72.94 \\ 
        all 50 & \textbf{93.34} & \textbf{39.38} & \textbf{91.52} & 82.10 & 74.37 & 72.63 \\ 
        all 100 & 93.21 & 40.56 & 91.32 & 82.13 & \textbf{72.99} & 72.68 \\ 
        only corr & 93.28 & 39.61 & 91.31 & 82.16 & 74.33 & \textbf{73.20} \\ 
        \hline
        & \multicolumn{6}{l|}{max logit} \\
        all 0 & 93.25 & 39.95 & 91.26 & 81.51 & 74.71 & 71.88 \\ 
        all 20 & 93.55 & \textbf{37.68} & 91.32 & \textbf{81.83} & 73.37 & 72.43 \\ 
        all 50 & \textbf{93.64} & 37.99 & \textbf{91.69} & 81.60 & 73.80 & 72.08 \\ 
        all 100 & 93.47 & 39.13 & 91.45 & 81.66 & \textbf{73.26} & 72.16 \\ 
        only corr & 93.60 & 37.98 & 91.46 & 81.66 & 73.92 & \textbf{72.63} \\ 
        \hline
        & \multicolumn{6}{l|}{energy} \\
        all 0 & 93.27 & 39.55 & 91.26 & 81.34 & 75.05 & 71.62 \\ 
        all 20 & 93.56 & \textbf{37.61} & 91.31 & \textbf{81.64} & \textbf{73.99} & 72.16 \\ 
        all 50 & \textbf{93.65} & 37.83 & \textbf{91.68} & 81.46 & 74.95 & 71.85 \\ 
        all 100 & 93.50 & 38.85 & 91.45 & 81.54 & 74.14 & 71.92 \\ 
        only corr & 93.62 & 37.75 & 91.45 & 81.49 & 74.69 & \textbf{72.37} \\ 
        \hline
        & \multicolumn{6}{l|}{React} \\
        all 0 & 93.28 & 39.55 & 91.26 & 81.34 & 75.05 & 71.63 \\ 
        all 20 & 93.56 & \textbf{37.61} & 91.31 & \textbf{81.65} & \textbf{73.99} & 72.17 \\ 
        all 50 & \textbf{93.65} & 37.83 & \textbf{91.68} & 81.47 & 74.95 & 71.86 \\ 
        all 100 & 93.50 & 38.85 & 91.45 & 81.54 & 74.14 & 71.93 \\ 
        only corr & 93.62 & 37.74 & 91.45 & 81.49 & 74.69 & \textbf{72.38} \\ 
        \hline
        & \multicolumn{6}{l|}{grad norm} \\
        all 0 & \textbf{90.85} & \textbf{48.28} & \textbf{88.76} & 68.25 & 84.71 & 58.46 \\ 
        all 20 & 89.48 & 51.01 & 87.28 & 68.61 & \textbf{83.80} & 59.26 \\ 
        all 50 & 89.11 & 50.96 & 87.19 & 68.70 & 85.39 & 59.15 \\ 
        all 100 & 89.12 & 50.90 & 87.13 & 68.66 & 84.26 & 59.01 \\ 
        only corr & 88.42 & 52.09 & 86.39 & \textbf{69.10} & 84.24 & \textbf{59.82} \\ 
        \hline
        & \multicolumn{6}{l|}{ViM} \\
        all 0 & 93.26 & 39.74 & 91.25 & 81.60 & 74.24 & 71.85 \\ 
        all 20 & 93.55 & \textbf{37.42} & 91.30 & \textbf{81.90} & 74.26 & 72.42 \\ 
        all 50 & \textbf{93.63} & 37.94 & \textbf{91.67} & 81.71 & 74.39 & 72.06 \\ 
        all 100 & 93.48 & 39.04 & 91.44 & 81.78 & \textbf{73.92} & 72.13 \\ 
        only corr & 93.60 & 37.74 & 91.44 & 81.70 & 74.43 & \textbf{72.59} \\ \hline
    \end{tabular}
\caption{Analysis of ALS variants. Only corr suggests using ALS only in correctly predicted training samples. Another strategy is to use all samples, increasing $\lambda$ from Epoch 0 to a predefined epoch such as 20 and 50. AUROC (m@1), FPR (m@2), and OSCR (m@3) are reported. The boldface shows the best for each case.}
\label{table5}
\end{table}

\begin{table}[!ht]
    \centering
    \begin{tabular}{|l|l|l|}
    \hline
        & CIFAR10 & TinyIN \\ \hline
        all 0 & 96.74 & 82.64 \\ 
        all 20 & 96.38 & 83.30 \\ 
        all 50 & \textbf{96.81} & 83.00 \\ 
        all 100 & 96.71 & 82.80 \\ 
        only corr & 96.67 & \textbf{83.46} \\ \hline
    \end{tabular}
\caption{Test accuracy in known data for ALS variants.}
\label{table6}
\end{table}

\textbf{Hyper-parameter $\lambda$}. As shown in Table \ref{table1}, test accuracy for our adaptive label smoothing is also improved and the optimal $\lambda$ partially depends on the dataset. Table \ref{table3} shows the average performance across the five datasets. With the most of OOD score functions, our method surpasses the baseline in AUROC, FPR, and OSCR. When $\lambda$ increases, the performance is firstly improved and then degrades. Although the optimal hyper-parameter varies for disparate score functions, it is not very sensitive and $\lambda=5$ successes for most cases.

\textbf{ALS strategy}. As mentioned in Section \ref{als}, our regularization for non-true class may be risky when the maximal probability does not give correct predictions. The variants is discussed here. We observed that the training is exploded when the regularization is used for all samples with a bigger $\lambda$ such as 100. In the main experiment, only the correct predictions are used with our regularization (only corr). Another strategy is to gradually improve $\lambda$ to a predefined value from the Epoch 0 until another settled epoch (all $T_e$): $\min \{e/T_e, 1\} * \lambda$ where $e$ is current epoch and $T_e$ is a hyper-parameter. We vary $T_e$ as 0, 20, 50, and 100 to see the difference. The results in CIFAR10 and TinyIN are shown in Table \ref{table5} with the test accuracy in known data given in Table \ref{table6}. Gradually increasing $\lambda$ gets the best test accuracy in CIFAR10 while the default strategy achieves the best in TinyIN. However, the variance of performance in CIFAR10 is smaller than that in TinyIN. We guess that classification in TinyIN is more difficult and more sensitive to the risk of ALS. Similar tendency is suggested in the OOD detection. The default strategy achieves the best OSCR for all score functions for TinyIN and another strategy is superior in CIFAR10. We argue that the optimal strategy depends on the complexity of a given application.

\begin{table}[!ht]
    \centering
    \begin{tabular}{|l|lll|lll|}
    \hline
        & m@1 & m@2 & m@3 & m@1 & m@2 & m@3 \\ 
        \hline
        & \multicolumn{3}{|c|}{max prob} & \multicolumn{3}{|c|}{energy} \\
        baseline & 87.09 & 55.88 & 82.55 & 89.27 & 46.42 & 83.74 \\ 
        LS 0.1 & 86.37 & 52.55 & 81.74 & 86.55 & 50.08 & 81.57 \\ 
        LS 0.2 & 86.13 & 52.46 & 81.57 & 86.27 & 50.26 & 81.35 \\ 
        LS 0.3 & 85.85 & 52.04 & 81.22 & 85.91 & 50.39 & 80.95 \\ 
        ALS 5 & 87.36 & 54.07 & 82.96 & 89.58 & 45.31 & 84.14 \\ 
        ALS 10 & 87.64 & 52.61 & 83.25 & \textbf{89.77} & \textbf{44.68} & 84.44 \\ 
        ALS 20 & 87.92 & 51.87 & 83.57 & \textbf{89.77} & 46.27 & 84.57 \\ 
        ALS 40 & 88.47 & \textbf{49.20} & \textbf{84.17} & 89.74 & 45.95 & \textbf{84.75} \\ 
        ALS 100 & \textbf{88.52} & 50.43 & 84.13 & 89.36 & 48.05 & 84.43 \\ 
        \hline
        & \multicolumn{3}{|c|}{entropy} & \multicolumn{3}{|c|}{React} \\
        baseline & 87.59 & 54.22 & 82.84 & 89.27 & 46.41 & 83.74 \\ 
        LS 0.1 & 86.68 & 50.88 & 81.83 & 86.53 & 50.08 & 81.54 \\ 
        LS 0.2 & 86.43 & 50.66 & 81.61 & 86.21 & 50.30 & 81.29 \\ 
        LS 0.3 & 86.08 & 50.85 & 81.20 & 85.79 & 50.40 & 80.84 \\ 
        ALS 5 & 87.86 & 52.97 & 83.27 & 89.59 & 45.31 & 84.14 \\ 
        ALS 10 & 88.21 & 51.94 & 83.62 & \textbf{89.77} & \textbf{44.68} & 84.44 \\ 
        ALS 20 & 88.74 & 51.08 & 84.11 & \textbf{89.77} & 46.26 & 84.57 \\ 
        ALS 40 & \textbf{89.17} & \textbf{48.65} & \textbf{84.64} & 89.74 & 45.94 & \textbf{84.75} \\ 
        ALS 100 & 88.92 & 49.89 & 84.39 & 89.36 & 48.04 & 84.43 \\ 
        \hline
        & \multicolumn{3}{|c|}{GEN} & \multicolumn{3}{|c|}{grad norm} \\
        baseline & 89.01 & 47.66 & 83.67 & 83.06 & 55.58 & 77.60 \\ 
        LS 0.1 & 86.60 & 50.38 & 81.64 & 84.57 & 53.77 & 79.59 \\ 
        LS 0.2 & 86.33 & 50.36 & 81.43 & 85.64 & 52.09 & 80.67 \\ 
        LS 0.3 & 85.97 & 50.43 & 81.03 & \textbf{86.19} & \textbf{51.05} & \textbf{81.17} \\ 
        ALS 5 & 89.27 & 46.94 & 84.05 & 83.79 & 54.74 & 78.35 \\ 
        ALS 10 & 89.49 & 46.18 & 84.35 & 84.14 & 53.54 & 78.84 \\ 
        ALS 20 & 89.68 & 46.87 & 84.61 & 84.22 & 55.17 & 79.04 \\ 
        ALS 40 & \textbf{89.79} & \textbf{45.90} & \textbf{84.88} & 85.09 & 53.03 & 80.12 \\ 
        ALS 100 & 89.46 & 47.83 & 84.59 & 85.70 & 53.32 & 80.74 \\ 
        \hline
        & \multicolumn{3}{|c|}{max logit} & \multicolumn{3}{|c|}{ViM} \\
        baseline & 89.26 & 46.60 & 83.80 & 89.29 & 46.40 & 83.77 \\ 
        LS 0.1 & 86.56 & 51.17 & 81.71 & 86.76 & 49.73 & 81.75 \\ 
        LS 0.2 & 86.26 & 51.39 & 81.53 & 86.46 & 50.21 & 81.54 \\ 
        LS 0.3 & 85.94 & 51.47 & 81.18 & 86.02 & 50.61 & 81.07 \\ 
        ALS 5 & 89.56 & 45.15 & 84.18 & 89.61 & 45.35 & 84.17 \\ 
        ALS 10 & \textbf{89.76} & \textbf{44.82} & 84.48 & \textbf{89.81} & \textbf{44.65} & 84.47 \\ 
        ALS 20 & \textbf{89.76} & 46.36 & 84.60 & \textbf{89.81} & 46.08 & 84.60 \\ 
        ALS 40 & 89.71 & 46.07 & \textbf{84.77} & 89.78 & 45.93 & \textbf{84.79} \\ 
        ALS 100 & 89.35 & 47.99 & 84.46 & 89.41 & 47.92 & 84.48 \\
        \hline
    \end{tabular}
\caption{Average performance on five datasets with light optimization, training with 200 epochs and mild data augmentation. m@1, m@2, and m@3 denote the average AUROC, FPR, and OSCR on five datasets.}
\label{table4}
\end{table}

\textbf{Light optimization}. We also probe the performance with less data augmentation and a small training epoch. Specifically, the models are trained with 200 epochs and the trained images undergo resizing, and random horizontal flip and gray. The experimental results are displayed in Table \ref{table4}. Again, our method benefits OOD detection. Label smoothing contributes to the classification of known samples but degrades the OOD detection. It suggests that, compared to more powerful optimization, stronger regularization for non-true classes is more beneficial to the lighter optimization, such that $\lambda=10$ tends to be decent. Similarly, performance tend to be improved and then reduced when $\lambda$ becomes bigger. In spite of optimal $\lambda$, the performance is not very sensitive. Although label smoothing secures the best result with grad norm score function, our method still has better result than the baseline for this specific score function. We also emphasize that grad norm can not achieve the best performance. In contrast, GEN, max logit, and ViM secure similar result.


\textbf{Additional dataset and backbone}. We also use our method in an additional dataset PaddyDoctor \cite{paddy_doctor} with 10,407 images in total, related to disease in paddy rice leaves. This dataset has ten classes, and six classes are randomly taken as known and four classes remained as unknown. Instead of VGG32 \cite{vgg}, ResNet50 \cite{resnet} is borrowed as the backbone. The optimization is the same as in TinyIN \cite{tinyimagenet} except for a smaller batch size 32. The experimental results are shown in Table \ref{table7}. Although LS surpasses the baseline in terms of test known accuracy with suitable hyper-parameter, LS receives much lower performance for OOD detection and ALS still secures the better performance. In addition, it is suggested that ALS outperforms LS using grad norm as the score function. In terms of ALS variants, ALS trained with all samples with increasing $\lambda=40$ until epoch 50 fails, and this variant performs worse than its counterparts with the same $\lambda$. We conclude that ALS with only correctly predicted samples with $\lambda=5$ almost achieves the best performance.

\begin{table*}[!ht]
\centering
\begin{tabular}{|l|l|l|l|l|l|l|l|l|l|l|l|l|}
\hline
& & \multirow{2}{*}{baseline} & \multicolumn{3}{c}{LS} & \multicolumn{4}{|c|}{ALS: only corr} & \multicolumn{3}{c|}{ALS: all 50} \\

\cline{1-2} \cline{4-13}
\multicolumn{2}{|l|}{hyper-parameter} & & 0.1 & 0.2 & 0.3 & 5 & 10 & 20 & 40 & 5 & 10 & 20 \\
\hline
\multicolumn{2}{|l|}{Acc of known} & 97.91 & \textbf{98.25} & 98.03 & 97.76 & 98.15 & 97.90 & 97.54 & 97.37 & 97.95 & 97.75 & 97.68 \\
\hline
max & AUROC & 92.38 & 81.16 & 81.24 & 78.15 & \textbf{92.54} & 92.30 & 91.14 & 91.08 & 92.01 & 91.15 & 89.12 \\
prob & FPR & 33.96 & 40.14 & 40.27 & 45.58 & \textbf{33.10} & 33.73 & 38.97 & 38.02 & 35.93 & 36.70 & 42.45 \\
 & OSCR & 91.38 & 80.29 & 80.27 & 77.23 & \textbf{91.56} & 91.31 & 90.03 & 89.95 & 91.05 & 90.17 & 88.14 \\
\hline
entropy & AUROC & 92.43 & 81.07 & 81.17 & 78.06 & \textbf{92.74} & 92.71 & 91.25 & 91.10 & 92.27 & 91.23 & 89.12 \\
 & FPR & 33.34 & 40.07 & 40.45 & 45.74 & \textbf{33.03} & 33.47 & 38.71 & 37.99 & 35.77 & 37.08 & 42.50 \\
 & OSCR & 91.42 & 80.19 & 80.19 & 77.13 & \textbf{91.78} & 91.74 & 90.12 & 89.96 & 91.28 & 90.19 & 88.14 \\
\hline
GEN & AUROC & 92.49 & 80.86 & 81.03 & 77.94 & \textbf{92.84} & 92.79 & 91.37 & 91.13 & 92.20 & 91.42 & 89.17 \\
 & FPR & 31.57 & 39.98 & 40.57 & 45.78 & \textbf{30.73} & 32.75 & 38.08 & 38.06 & 32.73 & 35.73 & 42.52 \\
 & OSCR & 91.44 & 79.97 & 80.05 & 77.00 & \textbf{91.84} & 91.78 & 90.23 & 89.99 & 91.20 & 90.36 & 88.18 \\
\hline
max & AUROC & 91.37 & 80.59 & 81.31 & 78.25 & 92.18 & \textbf{92.36} & 91.42 & 91.15 & 91.26 & 91.45 & 89.19 \\
logit & FPR & 34.47 & 40.16 & 40.63 & 45.68 & \textbf{31.79} & 34.25 & 38.05 & 38.11 & 34.22 & 35.00 & 42.61 \\
 & OSCR & 90.30 & 79.71 & 80.33 & 77.32 & 91.17 & \textbf{91.32} & 90.27 & 90.00 & 90.24 & 90.39 & 88.20 \\
\hline
energy & AUROC & 91.33 & 80.51 & 81.34 & 78.29 & 92.18 & \textbf{92.36} & 91.41 & 91.15 & 91.24 & 91.46 & 89.18 \\
  & FPR & 34.56 & 40.24 & 40.72 & 45.80 & \textbf{31.46} & 33.74 & 38.05 & 38.13 & 34.46 & 35.03 & 42.61 \\
 & OSCR & 90.26 & 79.62 & 80.36 & 77.35 & 91.16 & \textbf{91.31} & 90.26 & 90.01 & 90.22 & 90.39 & 88.20 \\
\hline
React & AUROC & 91.54 & 79.61 & 79.60 & 75.38 & \textbf{92.06} & 91.93 & 89.99 & 87.43 & 91.01 & 91.33 & 85.94 \\
 & FPR & 34.41 & 49.15 & 52.33 & 51.11 & \textbf{32.13} & 34.33 & 38.19 & 41.27 & 34.10 & 35.02 & 45.30 \\
 & OSCR & 90.46 & 78.74 & 78.60 & 74.44 & \textbf{91.05} & 90.88 & 88.85 & 86.30 & 90.00 & 90.25 & 84.98 \\
\hline
grad & AUROC & 72.97 & 76.17 & 77.31 & 76.75 & 70.49 & 69.60 & 77.73 & \textbf{83.92} & 70.94 & 68.61 & 73.68 \\
norm & FPR & 71.43 & 57.35 & 53.24 & 53.91 & 77.07 & 74.24 & 54.09 & \textbf{50.17} & 77.15 & 69.17 & 57.08 \\
 & OSCR & 71.97 & 75.26 & 76.35 & 75.77 & 69.61 & 68.65 & 76.46 & \textbf{82.68} & 70.02 & 67.58 & 72.66 \\
\hline
ViM & AUROC & 91.50 & 81.26 & 82.56 & 79.67 & 92.27 & \textbf{92.42} & 91.45 & 91.18 & 91.37 & 91.52 & 89.22 \\
 & FPR & 34.40 & 39.92 & 40.72 & 45.51 & \textbf{31.67} & 33.44 & 38.08 & 37.95 & 34.04 & 35.14 & 42.48 \\
 & OSCR & 90.42 & 80.36 & 81.57 & 78.71 & 91.26 & \textbf{91.37} & 90.30 & 90.04 & 90.36 & 90.45 & 88.24 \\
\hline
\end{tabular}
\caption{Performance in PaddyDoctor \cite{paddy_doctor}. We observed that ALS trained with all samples and increasing $\lambda=40$ until epoch 50 failed and their results are not given. The bold face shows the best for each case.}
\label{table7}
\end{table*}

\section{Conclusion}
In this paper, we first analyzed why label smoothing is not beneficial for OOD detection, in spite of its superiority in classifying known samples. The main reason is that label smoothing shrinks the maximal probability and logits, by which known and unknown samples have more overlap. We further rethought label smoothing from a disentangled perspective and found that the predefined limited target for known class results in the shrink and the same target for all non-true classes contributes to learn a better feature space. Inspired by this understanding, we proposed a new regularization, called adaptive label smoothing, to mitigate the issue in label smoothing, by punishing the difference between the non-maximal probabilities. The proposed method secured much better performance of OOD detection across five widely used public datasets and also showed its compatibility with different score functions, as well as light optimization. Our ALS also secured the best OOD detection performance in an additional dataset with ResNet50 as the backbone. We argue that our proposed method can be used as a plug-and-play method with other optimizations, taken as future work. Although we focus on OOD detection, we can imagine it for other tasks such as generic classification and knowledge distillation as future work.



\bibliographystyle{aaai}
\bibliography{aaai} 

\end{document}